\documentclass[twoside,11pt]{article}

\usepackage{fullpage}
\usepackage{dsfont}
\usepackage{comment}
\usepackage{textcomp}
\usepackage{amssymb}
\usepackage{bbm}
\usepackage{fancyhdr}
\usepackage{amsmath}
\usepackage{amsbsy,amsthm}
\usepackage{amscd}
\usepackage{latexsym}
\usepackage{graphicx}   %
\usepackage{pdfsync}
\usepackage{blkarray}
\usepackage{multirow}
\usepackage{hyperref}
\usepackage{xcolor}
\usepackage{enumerate}
\usepackage{booktabs}

\usepackage{tcolorbox}

\usepackage{tikz}

\usepackage{algorithmic}

\usepackage{natbib}

\newcommand{\R}{\mathbb{R}}

\newcommand{\N}{\mathbb{N}}

\newcommand{\E}{\mathbb{E}}

\newcommand{\cN}{{\cal N}}

\newtheorem{lemma}{Lemma}
\newtheorem{theorem}{Theorem}

\newtheorem{proposition}{Proposition}

\allowdisplaybreaks
\theoremstyle{remark}
\newtheorem{remark}{Remark}

\def\IND{\mathbbm{1}}

\newcommand{\ol}{\overline}
\newcommand{\wt}{\widetilde}
\newcommand{\wh}{\widehat}
\newcommand{\argmin}{\mathop{\mathrm{argmin}}}
\newcommand{\argmax}{\mathop{\mathrm{argmax}}}

\newcommand{\C}{\mathcal{C}}
\newcommand{\J}{\mathcal{J}}

\newcommand{\bj}{\boldsymbol{j}}
\newcommand{\bJ}{\boldsymbol{J}}
\newcommand{\bN}{\boldsymbol{N}}
\newcommand{\bmu}{\boldsymbol{\mu}}
\newcommand{\bq}{\boldsymbol{q}}

\newcommand{\EXP}{\mathbb{E}}
\newcommand{\PROB}{\mathbb{P}}

\newcommand{\defeq}{\stackrel{\mathrm{def.}}{=}}

\newcommand{\one}{\mathbbm{1}}

\newcommand{\cC}{\mathcal{C}} 
\newcommand{\cH}{\mathcal{H}}
\newcommand{\Reg}{\mathfrak{R}}

\newcommand{\UCB}{\textsc{UCB}}

\begin{document}

\title{{Bandit problems with fidelity rewards}}
\author{
G\'abor Lugosi
\thanks{Department of Economics and Business, Pompeu
Fabra University, Barcelona, Spain, gabor.lugosi@upf.edu}
\thanks{ICREA, Pg. Lluís Companys 23, 08010 Barcelona, Spain}
\thanks{Barcelona Graduate School of Economics}
\and 
Ciara Pike-Burke
\thanks{Department of Mathematics, Imperial College London, London, UK, c.pikeburke@gmail.com}
\and
Pierre-Andr\'{e} Savalle
\thanks{ Cisco Systems, Inc., Paris, France, psavalle@cisco.com}
}

\maketitle

\begin{abstract}%
 The \emph{fidelity bandits} problem is a variant of the $K$-armed
  bandit problem in which the reward of each arm is augmented by a
 fidelity reward that provides the player with an
  additional payoff depending on how ‘loyal’ the player has been to
  that arm in the past. We propose two models for fidelity. In the
  \emph{loyalty-points} model the amount of extra reward depends on
  the number of times the arm has previously been played. In the
  \emph{subscription} model the additional reward depends on the
  current number of consecutive draws of the arm.   We consider both 
  stochastic and adversarial problems. Since
  single-arm strategies are not always optimal in stochastic problems, the notion of regret in the
  adversarial setting needs careful adjustment. We introduce three possible notions of regret
  and investigate which can be bounded sublinearly.
  We study in detail the special cases of increasing, decreasing and coupon (where the player gets an
  additional reward after every $m$ plays of an arm) fidelity
  rewards.
  For the models which do
  not necessarily enjoy sublinear regret, we provide a worst case lower bound. For
  those models which exhibit sublinear regret, we provide algorithms and bound their regret. 
\end{abstract}

\section{Introduction}
Consider the problem of a worker searching for a good restaurant for their daily lunches. They are willing to explore the neighborhood in order to find the best restaurant, but also want to have good lunches. This is well modeled as a bandit problem, where the worker has to balance exploration (trying out new or unfamiliar restaurants), and exploitation (having a nice lunch at a favored place). However, the overall experience and satisfaction of the worker may not depend solely on the chosen restaurant, but also on whether the worker is a regular at this restaurant. 
Indeed, being loyal often leads to better experiences (e.g., waiters are more friendly, or the customer gets a free meal once in a while). We call the bonus from this loyalty the \emph{fidelity reward}.
In other situations, fidelity can have the opposite effect on the reward. For example, restaurants may offer free drinks to new customers, or customers may get bored of visiting the same restaurant. In this case the fidelity reward decreases. 

We consider multi-armed bandit problems where the reward of each arm is augmented by a {\it fidelity reward} that provides an additional payoff depending on how loyal the player has been to a given arm.
We consider two models for the fidelity rewards, namely the \emph{loyalty-points} model and the \emph{subscription} model.

Under the \emph{loyalty-points} model, the fidelity reward is a (possibly arm-specific) function of the total number of past draws of the arm. An important feature of these loyalty points is that once collected, they stay in the bank. This corresponds to the common practice of loyalty programs in marketing and retail, where the customer is rewarded for loyal behavior (e.g., the loyalty schemes offered by airline companies or grocery stores). 
However, in some cases, loyalty may cause the reward to decrease. For example
 in recommendation systems, users grow bored of repeatedly seeing the same content. In the stochastic bandit setting, the latter problem is known as \emph{rotting bandits} and has been studied by \citet{levine2017rotting}, \citet{seznec2018rotting,seznec2020single} and others. %
In this paper, we extend this to adversarial rewards and also consider fidelity functions that are not necessarily decreasing functions of the number of past plays. 

In the \emph{subscription} model, the fidelity reward is a (possibly arm-specific) function of the current number of consecutive draws of the arm. In particular, this means that if a player stops playing an arm, the next time they play it they will have to start the subscription again from scratch. This model is more realistic in situations where a user pays for a subscription, such as for a mobile phone plan or a gym membership. Here, the benefit of using a particular service increases with the length of continuous use. Again, there may be scenarios where continuously selecting the same option has a negative impact, for example when customers grow bored of going to the same restaurant every day, or where the benefit is only received after selecting an option a given number of consecutive times, such as when using a weekly metro pass.

The aim of this paper is to understand how the presence of fidelity rewards effects the difficulty of the multi-armed bandit problem. Our main interest is in the adversarial setting, where we assume that the base rewards are generated by some oblivious adversary and the player receives additional fidelity rewards depending on their past actions.
We wish to develop policies with low regret, %
defined as the cumulative difference of rewards between the policy and some baseline algorithms.
Clearly, the best baseline selects the best possible sequence of $T$ actions. However, even in the standard adversarial bandits problem without any additional fidelity rewards, it is impossible to compete with such a strong baseline. Instead, it is common to take inspiration from the stochastic setting, where it is known that an optimal policy is a single-arm policy that constantly plays the best arm, and compare the performance of an adversarial bandits algorithm to the best single-arm strategy. 
In the fidelity bandits problem, we also define the regret relative to a class of policies that is optimal in the stochastic problem. 
However, here the addition of fidelity rewards means that the optimal policy class in the stochastic setting may be more complex. Moreover, there may be multiple optimal policies for the stochastic problem, thus complicating the definition of regret in the adversarial setting. The first contribution of this paper is to define several natural definitions of regret for the adversarial fidelity bandits problem and to understand the relationship between them.

With these definitions of regret in hand, we aim to understand in which adversarial fidelity bandits problems is it possible to achieve regret that is sublinear in $T$. In the loyalty points problem, we show that when the fidelity rewards are increasing functions of the number of previous plays, sublinear regret is not possible for any definition of the regret that we consider. However, when the fidelity rewards are decreasing functions of the number of past plays, sublinear regret is possible for some definitions of regret in the loyalty points model. To show this, we introduce a bandit algorithm that exploits the concavity of the cumulative fidelity reward in this setting and achieves regret scaling with $K\sqrt{T}$. 
Lastly, in the so called `coupons' model where the fidelity reward is only received every $\rho_j$ plays of an arm, we show that it is possible to achieve sub-linear regret and provide an algorithm with regret scaling with $\sqrt{KT}$ in the loyalty points setting.
For the subscription model, where the fidelity rewards are functions of the number of consecutive plays of the arm, our results for the increasing case are the same as for the loyalty points model. Indeed, we show that there exist some settings with increasing fidelity functions where the regret must be linear. In the decreasing subscription model, the results are more positive. Indeed, here we are able to provide an algorithm whose strong regret is sublinear, and in particular is of order $(KT)^{2/3}$. We also show that sublinear regret is possible in the coupons model with subscription fidelity reward, although here the regret scales multiplicatively with the periodicity of the coupon function.
Our results are summarized in Tables~\ref{tab:summary} and \ref{tab:summarysub}. For completeness, we also provide a near-optimal algorithm for the stochastic fidelity bandits problem in the cases where no such algorithm exists in the literature.

\subsection{Related work}
In the loyalty points bandits problem, the reward changes depending on the past plays of each arm. Thus, the loyalty points bandits problem can be seen as an instance of the rested bandits problem introduced by \citet{whittle1988restless} which has been studied in the stochastic reward setting. We are not aware of any work on the rested bandits problem with adversarial rewards.
In the stochastic rested bandits problem, it is classically assumed 
that the reward of each arm is a stochastic process which only transitions to the next value when the arm is played. 
\citet{tekin2010online,tekin2012online} considered Markov processes. %
A more general process was considered by \citet{cortes2017discrepancy} but 
they considered weaker definitions of regret, and compared the reward of their algorithm to the best single arm strategy, or the best arm given the history of actions taken by their algorithm. %

The rotting bandits problem studied by \citet{levine2017rotting,seznec2018rotting,seznec2020single,bouneffouf2016multi} is another instance of the stochastic rested bandits problem. This problem is equivalent to the loyalty points bandit problem with stochastic base rewards and decreasing fidelity reward.
Indeed, here the expected reward decreases as a function of the number of plays of an arm.
The best result that we are aware of for this problem is from \citet{seznec2018rotting,seznec2020single} who show that the regret can be bounded by $\wt O(\sqrt{KT})$. For this, \citet{seznec2020single} present a natural UCB style algorithm (c.f. \citet{auer2002finite}) which does not use any knowledge of the fidelity functions.

In the subscription bandits problem, the reward at each time step depends on the previous plays of all arms. In particular, when we stop playing an arm after having played it consecutively for some number of steps, its reward changes even though it is not played, in addition to the reward of the played arm changing. This problem therefore does not exactly fit into either the classical rested or restless bandit framework, especially since while the restless bandit allows for the reward of an arm to change independently of whether it was played, this change is typically assumed to depend on the environment and not on the actions of the player, see \citet{whittle1988restless,slivkins2008adapting,garivier2011upper,besbes2014stochastic}.
We are not aware of any prior work studying this particular problem neither in the stochastic nor in the adversarial setting.

The stochastic bandit problem where the reward depends on the time between consecutive plays of an arm has been studied by \citet{immorlica2018recharging,pike2019recovering,cella2020stochastic,basu2019blocking}. 
This is potentially another form of measuring loyalty to an arm but is distinct to the model considered in this paper. 

Lastly, we observe that standard adversarial bandit algorithms are not sufficient to deal with the reward structure in the fidelity bandits problem.
These algorithms are typically designed to be competitive with the best single arm strategy which may be far from optimal in the fidelity bandit problem. Moreover, the addition of the fidelity rewards means that it is possible for the adversary to pick rewards to make either getting high reward or high fidelity reward impossible. Similarly, algorithms for learning in MDPs with adversarial rewards (e.g., \citet{even2009online,zimin2013online}) are not applicable. These require the MDP to be episodic or irreducible.

\section{Problem setup}
In the classical $K$-armed bandit problem, a player sequentially interacts with the environment. Let $[K]=\{1,\dots, K\}$ denote the set of arms. Then in each round, $t$, the player selects an arm $J_t \in [K]$, possibly using the previous observations and incorporating an element of randomness into their decision. The environment then returns a reward $X_{t,J_t}$ depending on which arm was selected. The player's aim is to select arms that maximize their total reward over $T$ rounds of the bandit game. 

In the \emph{stochastic} multi-armed bandit problem, the rewards $X_{t,j}$ for arm $j$ are generated i.i.d. from an underlying reward distribution. In particular, $\E[X_{t,j}] = \mu_j$ for all $1 \leq t \leq T$. We assume throughout the paper that the rewards are bounded in $[0,1]$. In the \emph{adversarial} multi-armed bandit problem,
we work under the model of an \emph{oblivious adversary}. In this model, 
all rewards $X_{t,j} \in [0,1]$ for $j\in [K]$ and $t\in [T]$ are chosen by an adversary prior to starting the game. 
At round $t \in [T]$ of the game, the player selects an arm $J_t \in [K]$ and the reward $X_{t,J_t}$ is revealed to the player, while 
the rewards $X_{t,j}$ with $j\neq J_t$ remain hidden forever. 

In this paper, we consider a modification of the standard multi-armed bandit problem to account for extra fidelity rewards. 
We focus on two models; the \emph{loyalty-points} model and the \emph{subscription} model. In both these models, after playing arm $J_t=j$ at time $t$, the player receives reward of the form,
\[ 
Y_{t,j} = X_{t,j} + \psi_j(\mathcal{H}_{t-1})~, 
\]
where $\psi_j(\mathcal{H}_{t-1})$ is some known, arm-specific function of the player's history $\mathcal{H}_{t-1} = \{ J_1,$ $Y_{1,J_1},$ $ X_{1,J_1} \dots, J_{t-1}, Y_{t-1,J_{t-1}}, X_{t-1,J_{t-1}} \}$. For ease of expression, we often refer to the $X_{t,j}$'s as the \emph{base rewards} and the $\psi_j(\mathcal{H}_{t-1})$'s as the \emph{fidelity rewards}. Therefore the reward $Y_{t,j}$ received at time $t$ is composed of the base reward from playing arm $j$ at time $t$ and the additional fidelity reward.

In the \emph{loyalty points} model, the reward depends on the number of past plays of each arm. Let $N_{t,j} = \sum_{s=1}^t \one_{\{J_{s} = j \}}$
denote the number of plays of arm $j$ up to time $t\ge 1$ and let
$N_{0,j}=0$.
(Here, and in the rest of the paper, $\one_{\{ \cdot \}}$ is used to denote the indicator function.)
Then, for all arms $j$, $\psi_j(\mathcal{H}_t) = f_j(N_{t,j})$ where each $f_j$ is a function defined on the set of natural numbers,
taking values in $[0,1]$.
Hence, the reward the player receives from playing arm $j$ at time $t$ is 
\[ 
Y_{t,j} = X_{t,j} + f_j (N_{t-1,j})~. 
\]

In the \emph{subscription} model, the fidelity reward depends on the current number of consecutive plays of the arm. Formally, define
\[
 Q_{t,j} = \one_{\{J_{t} = j \}} ( t - \max \{ s \leq t \,:\,J_s = j, J_{s-1} \neq j\})~.  
\]
Note that since only one arm can be played per time step, $Q_{t,j}=0$
for all arms $j \neq J_{t-1}$. 
In this case, $\psi_j(\mathcal{H}_{t-1}) = f_j(Q_{t,j})$ for some known, arm-specific function of the number of consecutive plays up to time $t-1$ and $f_j:\mathbb{N} \to [0,1]$ for all arms $j \in [K]$. The reward received from playing arm $j$ at time $t$ is then,
\[ 
Y_{t,j} = X_{t,j} + f_j (Q_{t,j})~.
\]

In this paper we assume that the fidelity functions $f_j$ are known to the player. 
In addition, we assume that the time horizon $T$ is also
known in advance. We also remark that in both models, when $\psi_j \equiv 0$ (so $f_j \equiv 0$) for all $j\in [K]$, the problem reduces to the classical multi-armed bandit problem.

\subsection{Regret}
\label{sec:regret}

We now turn to the problem of measuring the performance of an
algorithm in the fidelity bandits problem. Typically, the performance of bandit algorithms is measured in
terms of their regret, where the regret is the total
difference in reward accumulated by playing according to some optimal
policy and that from playing the algorithm of interest.  In order to
properly define meaningful notions of regret, we discuss the
stochastic and adversarial versions of the problems separately.

\subsubsection{Regret in the stochastic bandit problem}
\label{sec:stochregret}

Consider first the stochastic variant of the problem. Recall that here, the base reward (without the fidelity component), 
associated to arm $j$ at time $t$ is a random variable $X_{t,j}$, where the $X_{t,j}$ are independent random variables 
taking values in $[0,1]$, and $\EXP X_{t,j} = \mu_j$ for all $t\in [T]$ and $j\in [K]$.

A policy $\pi$ of the player is a sequence $(\pi_t)_{t=1}^T$ where $\pi_t$ is a mapping from the history of arms and rewards,
$\cH_{t-1} = \{\pi_1, Y_{1,\pi_1} , X_{1,\pi_1}\dots, \pi_{t-1}, Y_{t-1,\pi_{t-1}}, X_{t-1,\pi_{t-1}}\}$, 
to the set $[K]$ of arms. 
We write $\pi_t$ to denote the choice of arm made by the policy $\pi$ at time $t$. 
The history $\cH_t$ depends on the chosen arms $\pi_1,\ldots,\pi_t$. 
The fidelity reward depends on the history of actions chosen according to policy $\pi$ so we write $\cH_t(\pi)$ for the generated history under policy $\pi$ up to time $t$.

The regret of a policy is defined as the difference between the cumulative reward of the policy and that of the best
sequence of arm choices. To define the regret formally, let $[K]^T$ 
be the set of sequences of arms of the form $\bj =(j_1,\ldots,j_T)$ with $j_1,\ldots,j_T \in [K]$ for horizon $T \in \N$.  
We write 
$%
   S_T(\bj) = \sum_{t=1}^T Y_{t,j_t}
$ %
for the cumulative reward of the sequence $\bj$.
We then define the 
\emph{regret} 
of a candidate policy $\pi$ as 
\begin{align}
\label{eqn:regdef}
\max_{\bj \in [K]^T} S_T(\bj) - S_T(\pi)~.
 \end{align}

To avoid complications caused by random fluctuations, instead
of the regret defined above, it is
customary to consider the so-called \emph{pseudo-regret}
(see \citet{bubeck2012regret,LaSz20}).
In this notion, the base reward of each arm $X_{t,j}$ is replaced by
its expectation $\mu_j$.
Hence, the pseudo cumulative reward of policy $\pi$ is defined as
$\wt S_T(\pi)=\sum_{t=1}^T (\mu_{\pi_t} +  \psi_{\pi_t}(\cH_{t-1}(\pi)))$.
Similarly, for a competing reference sequence $\bj$, we write 
$\wt S_T(\bj) =\sum_{t=1}^T (\mu_{j_t} + \psi_{j_t}(\cH_{t-1}(\bj)))$.
We then define the pseudo-regret as
\begin{align}
\Reg_{T}(\pi)= \max_{\bj \in [K]^T} \wt S_T(\bj)- \wt S_T(\pi)~.\label{eqn:psreg}
 \end{align}
We often write $\Reg_{T} = \Reg_{T}(\pi)$ when the policy
$\pi$ is clear from the context.
Note that $\wt S_T(\pi)$ may still be random since the arm choices of
the policy $\pi$ %
influence the fidelity rewards. On the other hand, for each sequence
$\bj \in [K]^T$, the pseudo cumulative reward  $\wt S_T(\bj)$ is
deterministic.

Note that in the classical multi-armed bandit problem, the 
regret is usually defined with respect to
a much smaller
reference class, containing only \emph{single-arm strategies}.
These strategies are constant sequences of the form $\bj =
(j,j,\ldots,j)$ for $j\in [K]$. The two definitions are, in fact,
equivalent in this problem. To see this, simply notice that in the absence of fidelity
rewards, there is always a single-arm strategy that maximizes the
expected cumulative reward (i.e., $j^* \in \argmin_{j \in[K]}\mu_j$).%

In the presence of fidelity rewards, single-arm strategies
are not necessarily optimal. Depending on the nature of the fidelity
reward function $f_j$, one may still be able to determine a subclass $\J
\subset [K]^T$ that guarantees that, regardless of the distribution of
the base rewards (that is, regardless of the values of $\mu_1,\dots, \mu_K$), there always exists $\bj' \in \J$ such that 
$\wt S_T(\bj') =\max_{\bj \in [K]^T} \wt S_T(\bj)$. In other words, 
\[
   \Reg_{T}(\pi) = \sup_{\bj \in \J} \wt S_T(\bj)- \wt S_T(\pi)~.
\]
We call such a subset \emph{sufficient}.
In solving the regret minimization problem, it is important to
understand the sufficient subsets $\J$ that allow one such a
reduction. In particular, in each case it is helpful to identify 
\emph{minimal} sufficient sets of sequences. For a particular fidelity reward
$f$, a set $\J$ is minimal if for all proper subsets $\J' \subset
\J$ there exists a distribution of the arms such that 
$\max_{\bj \in \J'} \wt S_T(\bj)  < \max_{\bj \in \J} \wt
S_T(\bj)$. We define 
\[
\C_{f} = \{ \J \subset [K]^T: \J \text{ is minimal and sufficient}\}~
\]
to be the class of minimal sufficient sets for given fidelity functions $f$.
For example, for the classical multi-armed bandit problem 
when $f_j \equiv 0$, the class $\C_{f}$ of minimal
and sufficient sets contains just one set, namely the set
$\J_0=\{ (j,j,\ldots,j): j\in [K]\}$ of single-arm strategies.

\subsubsection{Regret in the adversarial bandit problem}

Next we discuss the possible notions of regret in the adversarial
setting. %
We work with an oblivious adversary. 
This means that the entire sequence of base rewards $(X_{t,j})_{t\in [T],j\in [K]}$ 
is fixed before play starts. In each round $t$, only the
reward corresponding to the arm chosen by the player is revealed.

In the adversarial setting the player is allowed to randomize the arm 
selection. (Without randomization, no nontrivial guarantees exist
even for the classical multi-armed bandit problem.) Formally, a
policy $\pi$ of the player is a sequence $(\pi_t)_{t=1}^T$ where $\pi_t$ is a mapping from the history of arms and rewards 
$\cH_{t-1} = \{J_1, Y_{1,J_1} , J_{1,J_1}\dots, J_{t-1}, Y_{t-1,J_{t-1}}, X_{t-1,J_{t-1}}\}$ 
to the set $\Delta_K$,  the standard simplex in $\R^K$. 
At time $t$, an arm $J_t$ is chosen at random, according to the
distribution $\pi_t(\cH_{t-1})$.

While in the stochastic multi-armed bandit problem the definition of
regret is quite natural, the adversarial case is significantly more complex. 
In the stochastic setting, we define the regret as the difference
between the (pseudo) cumulative reward of the player and that of the
best  sequence of actions. %
Adopting this notion to the adversarial framework often leads to trivialities. To see this, just consider 
the case of the classical adversarial multi-armed bandit problem where $f_j \equiv 0$ for all arms $j$.
In this case, the optimal policy is to play the arm with highest reward at each time step ($j_t \in \argmax_j X_{t,j}$). Hence, the 
player would need to compete with all possible $K^T$ sequences of arm choices. Obviously, achieving 
a sublinear regret with respect to this class of strategies is
hopeless. 

One usually gets around this 
problem by reducing the class of comparison policies
(\citet{auer1995gambling}). 
The usual definition of regret in the absence of fidelity rewards is with respect to 
the class of single-arm strategies.
This definition is natural since single-arm strategies are always
optimal in the classical stochastic bandit problem.
With this definition of regret, the adversarial multi-armed problem may be considered as a ``robust''
version of its stochastic counterpart. 

As explained in the previous sections, in the stochastic fidelity bandits problem, %
competing against the best single-arm strategy is often too
restrictive. %
It is clear that in many cases, it is
necessary to switch arms in order to get good reward. 
This means that the best single arm strategy is no longer a reasonable
baseline. 
Instead, we adopt the same philosophy as described above, and use knowledge of what makes a good sequence in the stochastic setting to define the regret in the adversarial case. 
In particular, to define the regret for the adversarial case---for given fidelity functions $f_j$---, we
first determine a minimal set of sequences of arm pulls that can be
optimal for the stochastic problem for some distribution. These are
the so-called sufficient and minimal sets defined in Section \ref{sec:stochregret}.

Then we define the regret as the difference of the cumulative reward of the player and that of the 
largest reward one could achieve by following one of the policies in a
sufficient and minimal comparison class. When there is only one
sufficient and minimal set (i.e., the class $\C_f$ contains a
single set $\J$), then the (pseudo) regret is defined simply as
\begin{equation}
\label{eq:regret}
   \Reg_T(\pi) = \max_{\bj \in \J} S_T(\bj)- \wt S_T(\pi)~,
\end{equation}
where $S_T(\bj)= \sum_{t=1}^T (X_{t,j_t} + \psi_{j_t}(\cH_{t-1}(\bj)))$ is the cumulative reward of a sequence
$\bj =(j_1,\ldots,j_T)$ and 
 $\wt S_T(\pi) = \EXP S_T(\bJ)$ is the expected cumulative reward of
 policy $\pi$. Here $\bJ=(J_1,\ldots,J_T)$ denotes the sequence of randomized 
arm choices of policy $\pi$ and the expectation is with respect to the
randomizations of the policy.

In some cases of fidelity rewards, there are multiple sets of minimal sufficient sets of
sequences. In such cases there are various natural ways of defining
regret. We single out two possibilities that we call the weak and strong regrets.
The \emph{weak regret} is defined as
\[
   \Reg_T^{\flat}(\pi) = \min_{\J \in \C_f} \max_{\bj \in \J} S_T(\bj)- \wt S_T(\pi)~,
\]
while the \emph{strong regret} is
\[
    \Reg_T^{\sharp}(\pi) =\max_{\J \in \C_f} \max_{\bj \in \J} S_T(\bj)- \wt S_T(\pi) 
   = \max_{\bj \in \bigcup_{\J \in \C_f} \J} S_T(\bj)- \wt S_T(\pi) ~.
\]
Clearly, minimizing the strong regret is a more ambitious goal, since
the player competes with a larger set of sequences of arms.
On the other hand, having a small weak regret means that the policy is
able to compete against at least one minimal sufficient set of
sequences
(whose identity may depend on the base rewards $(X_{t,j})$).
Of course, if $\C_f$ contains only one set, then the weak and
strong notions coincide and reduce to the definition \eqref{eq:regret}. In particular, note that when $f_j\equiv 0$, both the weak and strong regret correspond to the notion of regret classically considered in standard adversarial bandits problem.

\subsubsection{Mean regret for loyalty points bandits} \label{sec:meanreg}
For loyalty points bandits, there is a third notion of regret
that is perhaps more natural than the weak and strong regret. This is defined by observing that in the stochastic loyalty points model, the total reward only depends on the number of plays of each arm and the average base reward, and not on the order of plays. We may replicate this in the adversarial setting using empirical averages. We call this the \emph{mean regret}. Unfortunately, such a definition of regret is not suitable for the subscription model where the total reward inherently depends on the ordering, since the subscription model only offers fidelity reward for consecutive plays of an arm.

If a sequence $\bj =(j_1,\ldots,j_T) \in [K]^T$ plays arm $j$  
$N_{T,j}(\bj)=\sum_{t=1}^T \one_{\{j_t=j\}}$ times up to time $T$ %
then
in the stochastic analogue of loyalty points bandits problem, 
its (pseudo) reward is%
\[
\wt S_T(\bj)= \sum_{j=1}^K \left(\mu_jN_{T,j}(\bj)  + \sum_{n=1}^{N_{T,j}(\bj)} f_j(n) \right) = \sum_{j=1}^K  \mu_j\left(N_{T,j}(\bj) + F_j(N_{T,j}(\bj)) \right) 
\]
where $F_j(n) = \sum_{t=1}^n f_j(t)$ are the cumulative fidelity
rewards and $\mu_j$ is the expected value of the base reward of arm
$j$.
Hence, for given fidelity functions $f_1,\ldots,f_K$,
in the stochastic problem, $\wt S_T(\bj)$ %
only depends on the
number of times each arm is played and the expected base rewards.

To shorten notation, introduce the vector $\bN=(N_1,\ldots,N_K)$ of
nonnegative integers satisfying $\sum_{j=1}^K N_j=T$. $\bN$ is called
the \emph{type} of the sequence $\bj$ if $\sum_{t=1}^T \one_{\{j_t = j \}}=N_j$ for all $j\in [K]$.
Let $\bmu=(\mu_1,\ldots,\mu_K) \in [0,1]^K$ denote the
vector of means of the base rewards of the $K$ arms
and define,
\[
   \sigma(\bmu,\bN)   \defeq   
 \sum_{j=1}^K   \left(N_j \mu_j + F_j(N_j) \right)~.
\]
Then, we can write the cumulative reward of the optimal
sequence of actions as
$
\max_{\bj \in [K]^T} \wt S_T(\bj)  = \max_{\bN}    \sigma(\bmu,\bN)~
$
 and the regret for the stochastic problem as $\Reg_T(\pi) = \max_{\bN}   \sigma(\bmu,\bN) - \wt S_T(\pi)$.
We introduce some definitions to help
relate these types to
the minimal sufficient sets.
Fix some fidelity reward functions $f_1,\ldots,f_K$.
For a given $\bmu \in [0,1]^K$, let $\cN(\bmu)=\{\bN: \sigma(\bmu,\bN) =
\max_{\bN'} \sigma(\bmu,\bN')\}$ be the set of optimal types.
We call a set $B$ of types a \emph{covering set} if for all $\bmu \in
[0,1]^K$, $\cN(\mu) \cap B \neq \emptyset$. A covering set $B$ is
\emph{minimal} if no proper subset of $B$ is a covering set. 
Each minimal sufficient set of sequences $\J\in \C_f$ is such
that $\J$ contains exactly one sequence of each type of a minimal
covering set $B$.
We say that a type $\bN$ is \emph{admissible} if $\bN \in \bigcup_{\bmu \in
  [0,1]^K} \cN(\mu)$. %

In the adversarial bandit
problem, it is natural to replace the values $\mu_j$ by the
\emph{empirical averages}
\[
    \wh\mu_{T,j} = \frac{1}{T} \sum_{t=1}^T X_{j,t}~,
\]
and set  $\max_{\bN} \sigma(\wh\bmu_T,\bN)$ as a goal for the
learner, where   $\wh\bmu_T= (\wh\mu_{T,1},\ldots,\wh \mu_{T,K})$. With this in mind, we define the \emph{mean regret} as
\begin{eqnarray*}
   \Reg_T^{\natural}(\pi)& = & \max_{\bN} \sigma(\wh\bmu_T,\bN) -
   \wt S_T(\pi) %
   = \max_{\bN \in \N^K: \sum_{j \in [K]} N_j=T}   \sum_{j=1}^K
   \left(N_j \wh\mu_{T,j} + F_j(N_j) \right) - \wt S_T(\pi)~.
\end{eqnarray*}
Hence, the type $\bN$ maximizing $\sigma(\wh\bmu_T,\bN)$ may be
interpreted as the ``best response'' to the (unknown) empirical means
of the outcomes (in hindsight). This point of view is often adopted in
the theory of repeated games, see \citet{HaMa99,HaMa00,FoVo99}.
This gives the mean regret a natural interpretation that is a
generalization of most usual notions of regret.

The next proposition establishes the relationship between the weak, strong, and mean
regrets.

\begin{proposition}
\label{prop:threeregrets}
In the loyalty points model, regardless of the fidelity rewards and the
base rewards, for each policy $\pi$,
\[
      \Reg_T^{\flat}(\pi) \le \Reg_T^{\natural}(\pi) \le \Reg_T^{\sharp}(\pi)~.
\]
In particular, if there is only one minimal and sufficient set $\J$, then
all three regrets are equal to \eqref{eq:regret}.
\end{proposition}

\begin{proof}
The stated inequalities may be equivalently written as
\[
   \min_{\J \in \C_f} \max_{\bj \in \J} S_T(\bj) \le  \max_{\bN}
   \sigma(\wh\bmu_T,\bN)\le
   \max_{\J \in \C_f} \max_{\bj \in \J} S_T(\bj)~.
\]
Fix the fidelity reward functions $f_1,\ldots,f_K$.
Denote by $\ol\bN$ the type of the sequence that solves the min-max
optimization in the definition of the weak regret, that is, 
\[
  \min_{\J \in \C_f}
\max_{\bj \in \J} S_T(\bj)= \min_{\bj\in [T]^K: \bj
  \text{\ has type \ } \ol\bN} S_T(\bj)~.
\]
Observe that there is an admissible type with this property.

On the other hand, for any admissible type $\bN=(N_1,\ldots,N_K)$,
the average of the cumulative rewards over all sequences of type $\bN$ equals
\begin{eqnarray*}
    \frac{1}{\binom{T}{N_1,\ldots,N_K}} \sum_{\bj\in [T]^K: \bj
      \text{\ has type \ } \bN} S_T(\bj) & = &
 \frac{1}{\binom{T}{N_1,\ldots,N_K}} \sum_{\bj\in [T]^K: \bj
      \text{\ has type \ } \bN} \sum_{t=1}^T X_{t,j_t} + \sum_{j=1}^K
    F_j(N_j) \\
    & = &
    \sigma(\wh\bmu_T,\bN)~.
\end{eqnarray*}
Hence, 
\[
\min_{\J \in \C_f} \max_{\bj \in \J} S_T(\bj) \le
\sigma(\wh\bmu_T,\ol\bN)
\le  \max_{\bN}
   \sigma(\wh\bmu_T,\bN)~,
 \]
proving the first announced inequality. The second inequality is
proved similarly.
\end{proof}

We note that while the mean regret is not applicable in the
subscription model, 
$ \Reg_T^{\flat}(\pi) \le  \Reg_T^{\sharp}(\pi)$ obviously holds in
the subscription model as well. 
In fact, in all instances of the subscription model that we consider in this paper, it will turn out that there is only one minimal sufficient set, meaning that $ \Reg_T^{\flat}(\pi)  = \Reg_T^{\sharp}(\pi)$, and so there is no need to consider an intermediate definition of regret.

\begin{table}[t]
	\centering
	\begin{tabular}{c c c c}
	\toprule
							 & Increasing 					& Decreasing		 		& Coupons	\\
		\midrule
		\multirow{2}{*}{Stochastic}& $\E[\Reg_T] =   \wt O(K^{1/3} T^{2/3})$ & $\E[\Reg_T] = \wt O(\sqrt{KT})$ & $\E[\Reg_T] = \wt O(\sqrt{KT})$   \\
							 & {[ Section~\ref{sec:stocinc} ]} & {[ \citet{seznec2020single} ]}	& {[ Section~\ref{sec:stoccoup} ]} \\
		\midrule
		\multirow{2}{*}{Adversarial}&  $\begin{aligned}\E[\Reg_T^{\sharp}] &=\Omega(T) \\ \E[\Reg_T^{\flat}] &=\Omega(T) \\ \E[\Reg_T^{\natural}] &=\Omega(T) \end{aligned}$ 
								& $\begin{aligned} \E[\Reg_T^{\sharp}] &= \Omega(T) \\ \E[\Reg_T^{\flat}] &=\widetilde O(K\sqrt{T}) \\ \E[\Reg_T^{\natural}] &=\widetilde O(K\sqrt{T}) \end{aligned}$
								& $\begin{aligned} \E[\Reg_T^{\sharp}] &=\widetilde O(\sqrt{KT} + \ol{\rho}) \\ \E[\Reg_T^{\flat}] &=\widetilde O(\sqrt{KT}) \\ \E[\Reg_T^{\natural}] &=\widetilde O(\sqrt{KT}) \end{aligned}$ \\
							 & 	{[ Section~\ref{sec:advinc} ]}	&{[ Section~\ref{sec:advdec} ]}& {[ Section~\ref{sec:advcoup} ]}
	\\ \bottomrule
	\end{tabular} 
\caption{Summary of our main results in the loyalty points bandits problems. $\widetilde O$ indicates order up to log factors, and $\ol{\rho}$ is the least common multiple of the periodicity $\rho_1,\dots, \rho_K$ in the coupons reward model.}
\label{tab:summary}
\end{table}

\begin{table}[t]
	\centering
	\begin{tabular}{c c c c}
	\toprule
							 & Increasing 					& Decreasing		 		& Coupons	\\
		\midrule
		\multirow{2}{*}{Stochastic}& $\E[\Reg_T] =
                                             \wt O(T^{2/3}K^{1/3})$ & $\E[\Reg_T] =   \wt O(T^{2/3}K^{1/3})$ & $\E[\Reg_T] = \wt O(\sqrt{KT} )$   \\
							 & [ Section~\ref{sec:stocincsub} ] & [ Section~\ref{sec:stocdecsub} ]	& [ Section~\ref{sec:stoccoupsub} ] \\
		\midrule
		\multirow{2}{*}{Adversarial}& $\begin{aligned} \E[\Reg_T^{\sharp}] &=  \Omega(T) \\  \E[\Reg_T^{\flat}] &=   \Omega(T)\end{aligned}$
				&  $\begin{aligned} \E[\Reg_T^{\sharp}] &=  \wt O((KT)^{2/3}) \\ \E[\Reg_T^{\flat}] &=\wt O((KT)^{2/3})  \end{aligned}$
								& $\begin{aligned} \E[\Reg_T^{\sharp}] &= O(\sqrt{\ol{\rho} KT}) \\ \E[\Reg_T^{\flat}] &= O( \sqrt{\ol{\rho} KT}) \\  \end{aligned}$ \\
							 & 	[ Section~\ref{sec:advincsub}\footnotemark ]		&	[ Section~\ref{sec:advdecsub} ]	& [ Section~\ref{sec:advcoupsub} ]
	\\ \bottomrule
	\end{tabular} 
\caption{Summary of our main results in the subscription bandits problems. $\widetilde O$ indicates order up to log factors, and $\ol{\rho}$ is the least common multiple of the periodicity $\rho_1,\dots, \rho_K$ in the coupons reward model.}
\label{tab:summarysub}
\end{table}

 \footnotetext{This is the general result. For the special case where the fidelity functions are step functions with a step from 0 to 1 at $m$ we also show that the weak and strong regret can be bounded by $\wt O(T^{2/3}(Km)^{1/3})$.}
\subsection{Fidelity reward functions}

Throughout this work, we assume that the fidelity functions are known and $f_j$ takes values in $[0,1]$ for all arms $j \in [K]$. %
We consider both increasing and decreasing fidelity functions, in addition to a certain class of periodic fidelity rewards that we call the ``coupons'' model. %
In the \emph{increasing} rewards model, we assume that for all $j  \in
[K]$, $f_j(n)$ is a nondecreasing function of $n$. For the
\emph{decreasing} rewards model, we assume that for all $j  \in [K]$,
$f_j(n)$ is nonincreasing in $n$. In the \emph{coupons model}, a
fidelity reward is obtained only once every few plays of each arm. In
particular, let the period length of arm $j$ be denoted by $\rho_j\in
\N$, and the bonus fidelity reward $r_j\in [0,1]$. Then, the player
receives a fidelity bonus $r_j$ after every $\rho_j$ plays of arm $j$. More formally, for $r_j>0$, 
 \[ f_j(t) = \begin{cases}
			r_j & \text{ if } t = 0 \pmod{\rho_j} %
			\\0 & \text{ otherwise.}
 		\end{cases}
 \]
 
Throughout the paper, it is often helpful to consider the cumulative reward of each arm $j$ which is defined for any $n \in [T]$ as,
\[ F_j(n) = \sum_{t=1}^n f_j(n)~. \]

We now consider the different cases in turn. For each, we first
analyze the stochastic setting and define the class of minimal
sufficient sequences which also defines the regret in the adversarial
setting. We then analyze the adversarial setting. Our results are
summarized in Tables~\ref{tab:summary} and \ref{tab:summarysub}.

\section{Loyalty points model: increasing fidelity rewards}
The first setting we consider is the loyalty points model where the fidelity functions are increasing functions of the number of past plays of each arm.

\subsection{Minimal sufficient sets}
We first need to define
the minimal sufficient sets $\J$ of sequences of actions defined in
Section \ref{sec:regret}.
When the fidelity rewards are nondecreasing, the unique minimal
sufficient set of sequences is the set $\J_0$ of single-arm
strategies. 
This is shown in the below lemma, whose proof is in Appendix~\ref{app:loyincopt}.
When the fidelity functions of the different arms are
equal, optimality of such a single-arm strategy is obvious. 
In the case of arm-specific fidelity functions, recall that $F_j(T) = \sum_{t=1}^T f_j(t)$, then the optimal arm is given by
\[
j^* = \argmax_{1 \leq j \leq K} \left\{\mu_j + \frac{1}{T} F_j(T) \right\}~.
\]
(In case of multiple maximizers, we may choose any of them). The optimal policy then plays arm $j^*$ for all rounds $t=1,\dots, T$.

\begin{lemma}
\label{lem:loyinc}
In the stochastic loyalty points model with increasing fidelity
rewards, regardless of the distribution of the base rewards and the
fidelity rewards, there exists a  single-arm strategy that minimizes
the pseudo cumulative reward $\wt S_T(\bj)$ over all $\bj \in [K]^T$.
\end{lemma}

\subsection{Stochastic rewards} \label{sec:stocinc}

As Lemma \ref{lem:loyinc} shows,
to minimize regret, it suffices to construct a strategy that is able to
compete with single-arm strategies. 
In the stochastic setting, a simple variant of the UCB algorithm of
\citet{auer2002finite} achieves this. 
Instead of using the observations $Y_{t,j}$ to construct the
sample averages and upper confidence bounds, one may work with the modified rewards 
$%
\widetilde Y_{t,j} = X_{t,j} + \frac{1}{T} F_j(T)~.
$ %
Note that since the
fidelity functions are known, computing these modified rewards is possible. 
These modified rewards are natural since the best
single-arm policy equates to playing the arm that maximizes $\mu_j +
F_j(T)/T$ in all rounds.
We then define the upper confidence bounds around the sample mean of
these observations by

\begin{align}
\label{eqn:cbslpi}
 \UCB_t(j)=\ol{X}_{t,j} + \frac{1}{T}  F_j(T) + \sqrt{ \frac{ 2\log(KT)}{N_{j,t}}}~,
\end{align}
where $\ol X_{t,j} = \frac{1}{N_{t,j}} \sum_{s=1}^t X_{s,j} \one_{\{
  J_s=j\}}$. %
The algorithm proceeds as a standard UCB algorithm: first it plays each arm once, then for $t=K+1,\dots, T$, it plays arm
\[ J_t= \argmax_{1 \leq j \leq K} \UCB_{t-1}(j)~. \]
The regret of this algorithm is bounded in the next theorem, whose proof follows from a modification of the standard UCB analysis (see \citep{auer2002finite}) and is given in Appendix~\ref{app:regstocsubinc}.

\begin{theorem}
\label{thm:regstocsubinc}
The expected (pseudo) regret of the UCB algorithm defined above
in the stochastic loyalty points bandits problem with increasing fidelity reward
is bounded by
\[ 
\E \Reg_T \leq \sum_{j:j \neq j^*}^K \frac{16 \log(TK)}{\wt \Delta_j^2}  (\mu_{j^*} - \mu_j + f_{j^*}(T)-f_j(0)) + \frac{1}{K}~,
\]
where $\widetilde \Delta_j = \mu_{j^*} - \mu_j + (F_{j^*}(T) - F_j(T))/T$.
Consequently, the worst case regret of this algorithm is bounded by $O(T^{2/3} (K\log(T))^{1/3})$.
\end{theorem}

When the fidelity functions are all constant, $f_{j^*}(T) =F_{j^*}(T)/T$, $ f_j(0) = F_j(T)/T$, the problem reduces to the standard stochastic multi-armed bandit problem
 and we recover the optimal $O(\sum_{j=1}^K \log(T)/\wt \Delta_j)$ order of the problem dependent regret (see \citet{auer2002finite}). 
When the fidelity functions are not constant, the leading term in the
regret is multiplied by $\frac{\mu_{j^*} - \mu_j +f_{j^*}(T) -
  f_j(0)}{\wt \Delta_j} \geq 1$. This term can be as large as $2/\wt \Delta_j$, in which case, 
the regret is of the order of $\frac{\log(KT)}{\wt \Delta_j^2}$. 
Thus, while the logarithmic dependence on the horizon $T$ is the same as in the standard multi-armed bandit problem,
the dependence on the gap $\wt \Delta_j$ may be worse, depending on the fidelity functions. 

We argue that this worse dependence is inevitable. 
Consider the case where the fidelity functions are the same for all arms, that is, $f_j=f$ for all $j\in [K]$. %
Here, $\wt \Delta_j = \mu_{j^*} - \mu_j$ and $j^*= \argmax_j \mu_j$. %
For simplicity, assume that the distribution of arm $j$ is Bernoulli with parameter $\mu_j\in [0,1]$.
If $F_j(T)=F_{j^*}(T)$, then to distinguish $\mu_j + F_j(T)/T$ from $\mu^* +F_{j^*}(T)/T$, if suffices to distinguish $\mu_j$ from $\mu_{j^*}$.
A classical result of \citet{lai1985asymptotically} shows that, in the standard multi-armed bandit problem, 
any algorithm that is guaranteed to achieve regret $o(T^\alpha)$ for every $\alpha>0$ 
must play any arm $j$ at least $N_j = \Omega(\frac{\log(T)}{\Delta_j^2})$ times to distinguish $\mu_j$ and $\mu_{j^*}$. 
Thus, any algorithm that obtains sublinear regret must explore sub-optimal arms which necessarily results in some loss of fidelity reward.
Define $\tau = \sum_{j: j \neq j^*} N_j $. Then the contribution to the regret from the lost fidelity reward is %

\begin{align*}
 \sum_{t=1}^T f(t) - \sum_{t=1}^T f(N_{t-1,J_t}) &= \sum_{t=T-\tau}^T f(t) - \sum_{j: j \neq j^*} \sum_{t=1}^{N_j} f(t) %
\geq \sum_{j: j \neq j^*} \sum_{t=1}^{N_j}(f(T-\tau)-f(N_j)) %
 \end{align*}
This 
gives a lower bound on the regret of $\Omega(\sum_{j: j \neq j^*}\frac{\log(T)}{\wt \Delta_j^2} (\mu_{j^*} - \mu_j+ f(T-\tau) - f(N_j)))$. 
While this does not exactly match %
Theorem~\ref{thm:regstocsubinc}, it shows that a term involving the fidelity functions is unavoidable and that
the problem is, in general, harder than the plain stochastic bandit problem. 

\subsection{Adversarial rewards} \label{sec:advinc}

As seen in the previous section, the only set of minimal sufficient
sequences is the set $\J_0$ of single-arm strategies. Hence, as in \eqref{eq:regret}, the 
notions of weak, strong, and mean regrets introduced in Section
\ref{sec:regret} coincide and equal 
\[
   \Reg_T(\pi) = \max_{\bj \in \J_0} S_T(\bj)- \wt S_T(\pi)~.
\]

This is the same notion of regret as in the standard adversarial
bandits problem.
In spite of the
simplicity of the target and unlike in the stochastic version of the
problem, here the addition of the fidelity reward makes it
considerably more difficult to achieve sublinear regret. In fact,
Theorem~\ref{thm:TregLoy} below shows that, unless the fidelity
rewards are essentially constant, 
any policy must incur linear regret for some reward sequence. 

To prove the lower bound, it suffices to consider the case where the fidelity function is the same for all arms. %
We obtain the following lower bound for the regret, the proof of which is in Appendix~\ref{app:loyinclb}.

\begin{theorem} 
\label{thm:TregLoy}
Consider the adversarial loyalty points bandits problem with increasing fidelity rewards, where the fidelity
function is the same for every arm. Define $\delta = f(7T/8)-f(T/8)$.
Then for every policy $\pi$ there exists a sequence of rewards such that the regret
satisfies
\[ 
 \Reg_T(\pi) \ge \frac{T\delta}{40}~. 
\]
\end{theorem}

Thus,  the regret is $\Omega(T)$ unless $f$
is essentially constant over most of its domain. The particular constants appearing in the bound (and in the definition of $\delta$) have no special significance, they have been chosen for convenience. In particular, the definition of $\delta$
may be replaced by $f(T(1-\epsilon))-f(T\epsilon)$ for any $\epsilon
\in (0,1/2)$ and the constant $1/40$ modified accordingly.
 We omit the straightforward details.

Theorem \ref{thm:TregLoy} implies that in the adversarial case one cannot hope for a nontrivial regret bound unless the (nondecreasing) fidelity
rewards essentially do not change with time. In that case one may use a simple and natural modification of 
any low-regret algorithm for adversarial bandits such as EXP3 (\citet{auer2002nonstochastic}). We omit the straightforward details.

\section{Loyalty points model: decreasing fidelity rewards}

\subsection{Stochastic rewards}

The stochastic loyalty points bandit problem with decreasing
fidelity rewards is equivalent to the \emph{rotting bandits}
problem studied by
\citet{levine2017rotting,seznec2018rotting,seznec2020single}, and
\citet{bouneffouf2016multi}. \citet{seznec2018rotting,seznec2020single}
provide an algorithm with regret
$\Reg_{T} =\wt O(\sqrt{KT\log(T)})$ which matches the lower bound for the standard bandit problem up to logarithmic factors.

In the adversarial setting, the loyalty points model with decreasing
fidelity rewards is significantly more complex. 
In order to appropriately define the notion of regret, first we need
to determine the class $\C_f$ of minimal sufficient sets of sequences of arm
pulls. 
We discuss this in the next section. 
As it turns out, the class $\C_f$ is nontrivial in this case and
therefore the notions of strong, weak, and mean regret do not coincide. In
fact, we show that while there is no algorithm that achieves sublinear 
strong regret for a large class of fidelity functions, it is
possible to construct a policy with guaranteed sublinear mean regret and thus also sublinear weak regret.

\subsection{Minimal sufficient sets}

In order to study the adversarial version of the problem, first we need to 
understand  the minimal sufficient sets of sequences of
actions. Recall from Section \ref{sec:regret} that each minimal sufficient set $\J\in \C_f$ is such that
$\J$ contains exactly one sequence of each type of a minimal
covering set $B$.
Hence, in order to understand the class $\C_f$ of minimal
sufficient sets, one needs to understand the set of admissible types, that is, types that are maximizers of
the function $\sigma(\bmu,\bN)$ for some $\bmu \in [0,1]^K$.

An important special case is when the fidelity rewards $f_j(t)$ are
\emph{strictly} decreasing functions of $t$ for each $j \in
[K]$. Then the average cumulative reward associated to a 
a type $\bN=(N_1,\ldots,N_K)$
\[
    h\left(\frac{\bN}{T}\right)= \frac{1}{T} \sum_{j=1}^K F_j(N_j)
  \]
is strictly concave. In fact,   
one may extend $h$ to the standard simplex $\Delta_K$
such that the function $h: \Delta_K \to \R$ is strictly concave.
Then the cumulative total expected reward corresponding to a mean
vector $\bmu \in [0,1]^K$ may be written as
\[
   \sigma(\bmu,\bN) = T \left(  \left\langle \frac{\bN}{T}, \bmu
     \right\rangle + h\left(\frac{\bN}{T}\right) \right)~.
 \]
It is easy to see that for such fidelity reward functions, for each
type $\bN$, there exists $\bmu \in [0,1]^K$ such that $\cN(\bmu)=
\{\bN\}$,
that is, $\bN$ is the unique type maximizing $\sigma(\bmu,\cdot)$.

In such cases, the minimal sufficient sets
are all those sets $\J$ that, for each type $\bN$, $\J$ 
contains exactly one sequence $\bj$ of type $\bN$.
But then
\[
\max_{\J \in \C_f} \max_{\bj \in \J} S_T(\bj) = \max_{\bj \in [K]^T} S_T(\bj) 
\]
and therefore it is hopeless to minimize the strong regret. 
Based on this, it is not difficult to prove the following proposition.

\begin{proposition} \label{prop:advdec}
Consider the adversarial loyalty points bandit problem
with strictly decreasing fidelity reward functions $f_1,\ldots,f_K$.
Then there exists a positive constant $c$ -- depending on the fidelity
reward functions only -- such that for any (randomized) policy $\pi$ of the forecaster, there exist 
base rewards $(X_{t,j})_{t\in [T],j\in [K]}$ such that the strong regret satisfies
\[
     \Reg_T^{\sharp}(\pi) \ge c T~.
\]
\end{proposition}

\begin{remark}
In some cases, when the fidelity rewards are not strictly decreasing (so the conditions of Proposition~\ref{prop:advdec} do not hold),
it is possible to achieve sublinear strong regret. As an example,
consider the case when $K=2$ and the fidelity function of both
arms is $f_j(t) = \alpha \one_{t \le T_0}$ for $j=1,2$, where
$\alpha \in (0,1)$ and $T_0 \in [T]$ are some fixed parameters.
Then it is easy to see that the set of optimal types $\cN(\bmu)$
corresponds
to a single-arm strategy whenever $|\mu_1-\mu_2| \neq \alpha$. When
$\mu_1-\mu_2= \alpha$, $\cN(\bmu)$ consists of all types with $N_1 \ge
T- T_0$ and when
$\mu_1-\mu_2= - \alpha$, $\cN(\bmu)$ consists of all types with $N_1
\le T_0$. Hence, in all cases, $\cN(\bmu)$ contains one of the two
single-arm strategies and therefore the only minimal sufficient set is
$\J_0$. This implies that the three notions of regret coincide in this
case.  As is shown below, sublinear mean regret is always
achievable and hence, in this particular example, the strong regret
can also be made sublinear.
\end{remark}

\subsection{Minimizing the mean regret} \label{sec:advdec}

Interestingly, as opposed to the strong regret, minimizing the mean
(and therefore weak) regret is an achievable goal. In this section we 
define a policy that guarantees that if the fidelity functions are
nonincreasing, then the mean regret is 
$o(T)$ regardless of the base rewards.

\begin{figure}
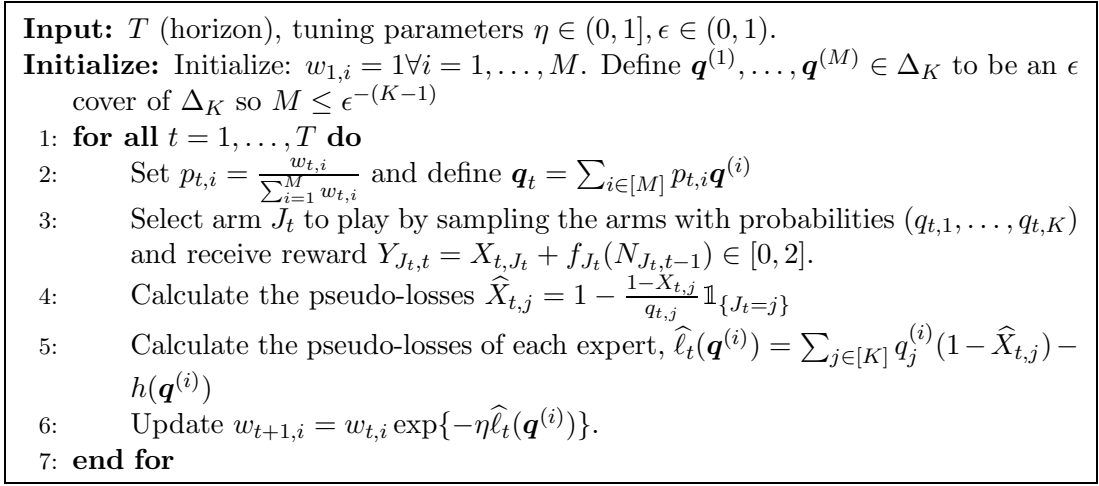

 \centering
\fbox{ \parbox{0.85\textwidth}{
\begin{algorithmic}[1]
\renewcommand{\algorithmicrequire}{\textbf{Input:}}
\renewcommand{\algorithmicensure}{\textbf{Initialize:}}
 \algsetup{indent=2em}
\REQUIRE $T$ (horizon), tuning parameters $\eta \in (0,1], \epsilon \in (0,1)$.
\ENSURE Initialize: $w_{1,i} = 1 \forall i=1, \dots, M$. Define $\bq^{(1)}, \dots, \bq^{(M)} \in \Delta_K$ to be an $\epsilon$ cover of $\Delta_K$ so $M\leq \epsilon^{-(K-1)}$
\FORALL{$t=1,\ldots,T$}
 	\STATE Set $p_{t,i} =\frac{w_{t,i}}{\sum_{i=1}^M w_{t,i}}$ and define $\bq_t = \sum_{i \in [M]} p_{t,i} \bq^{(i)}$
 	\STATE Select arm $J_t$ to play by sampling the arms with probabilities $(q_{t,1}, \dots, q_{t,K})$ and receive reward $Y_{J_t,t} = X_{t,J_t} + f_{J_t}(N_{J_t,t-1}) \in [0,2]$.
 	\STATE Calculate the pseudo-losses $\wh X_{t,j} = 1-\frac{1- X_{t,j}}{q_{t,j}} \one_{\{J_t=j\}}$ 
 	\STATE Calculate the pseudo-losses of each expert, $ \wh\ell_t(\bq^{(i)})=   \sum_{j\in [K]} q^{(i)}_j (1-\wh{X}_{t,j}) -
 h(\bq^{(i)})$
 	\STATE Update $w_{t+1,i} = w_{t,i} \exp \{-\eta \wh \ell_t(\bq^{(i)})\}$.
 \ENDFOR
 \end{algorithmic}
 }
 }
\caption{EXP4 with pseudo rewards for the adversarial loyalty points bandits model with decreasing fidelity reward.}
\label{alg:exp4}
\end{figure}

The algorithm we propose is an exponentially weighted average
forecaster where the set of experts $\bq^{(1)}, \dots, \bq^{(M)} \in
\Delta_K$ is defined by an $\epsilon$-cover of the simplex
$\Delta_K$.
Similarly to in the standard exponentially weighted average forecaster, we need to estimate the loss of each expert. In this loyalty points bandits problem this is somewhat more involved since the loss estimates need to be defined to take into account the fidelity rewards. 
We use the loss estimates $ \wh\ell_t(\bq^{(i)})=   \sum_{j\in [K]} q^{(i)}_j (1-\wh{X}_{t,j}) -
 h(\bq^{(i)})$ for each expert $i$ where $\wh X_{t,j} = 1-\frac{1- X_{t,j}}{q_{t,j}} \one_{\{J_t=j\}}$ for all arms $j$.
 The algorithm is defined in Figure~\ref{alg:exp4}.
 In order to work with these adjusted losses, we need to show that
 they enjoy the desirable properties of an importance sampling
 estimate and that they lead to the correct dependence on the fidelity reward of our algorithm.
 This is done in the following theorem which provides a bound on the regret of the algorithm.

\begin{theorem} \label{thm:loydecreg}
Consider the adversarial loyalty points bandit problem when the fidelity rewards are nonincreasing for all $j\in [K]$. 
Then, regardless of the base rewards, the randomized policy $\pi$ defined in Figure~\ref{alg:exp4} with $\epsilon = (K+1)/\sqrt{2T}$ and $\eta = \sqrt{\log(1/\epsilon)/(2T)}$ has regret
\[
  \EXP \Reg_T^{\natural}(\pi) \le (K+1)\sqrt{2T} \left(\frac{1}{2}
  \sqrt{\log(K+1)} + 1+ \sqrt{\log\frac{\sqrt{2T}}{K+1}} \right)~.
\]
\end{theorem}

\begin{proof}
We consider a randomized forecaster that, at each time instance, 
selects a point $\bq_t=(q_{t,1},\ldots,q_{t,K})$  in 
the standard simplex $\Delta_K$. Then the forecaster plays a random
arm, chosen according to the distribution $\bq_t$, that is, 
\[
    J_t = j \quad \text{with probability $q_{t,j}$,} \qquad j \in [K]~,
\]
Writing $\wt{N}_j= \sum_{t=1}^T \one_{J_t=j}$ and
$\wt\bN=(\wt{N}_1,\ldots,\wt{N}_K)$ for the corresponding type,
the cumulative reward of such a forecaster
equals
\[
   \sum_{t=1}^T X_{t,J_t}+ \sum_{t=1}^T F_j(\wt{N}_j) = T \left(
     \frac{1}{T} \sum_{t=1}^T X_{t,J_t} +
     h\left(\frac{\wt\bN}{T}\right) \right)~,
 \]
 Consider the piecewise linear extension of $h$ to $\Delta_K$.
 It follows from the fact that the fidelity rewards are nonincreasing
 that
 $h$ is concave and $1$-Lipschitz with respect to the $\ell_1$
norm. Then we may apply the Hoeffding-Azuma inequality to both
$\sum_{t=1}^T X_{t,J_t}$ and $\wt{N}_j$ to obtain that, with
probability at least $1-\delta$, 
\begin{eqnarray*}
  \sum_{t=1}^T X_{t,J_t}+ \sum_{t=1}^T F_j(\wt{N}_j)
  & \ge &
         \sum_{t=1}^T \sum_{j\in [K]} q_{t,j} X_{t,j} +
    T h\left(\frac{1}{T} \sum_{t=1}^T \bq_t\right)  -
          (K+1)\sqrt{\frac{T}{2}\log\frac{K+1}{\delta}}  \\
  & \ge &
          \sum_{t=1}^T \sum_{j\in [K]} q_{t,j} X_{t,j} +
    \sum_{t=1}^T  h\left(\bq_t\right)- (K+1)\sqrt{\frac{T}{2}\log\frac{K+1}{\delta}}~,
\end{eqnarray*}
where we used concavity of $h$.
In particular, by standard tail integration, 
\begin{eqnarray}
\label{eq:azuma}
\lefteqn{   \EXP  \left[ \sum_{t=1}^T X_{t,J_t}+ \sum_{t=1}^T F_j(\wt{N}_j) \right]
  } \nonumber \\
  & \ge &
       \EXP  \left[ \sum_{t=1}^T \sum_{j\in [K]} q_{t,j} X_{t,j} +
    \sum_{t=1}^T  h\left(\bq_t\right)  \right]  - (K+1)\sqrt{\frac{T}{2}\log(K+1)}~,
\end{eqnarray}
Hence, it suffices to construct $\bq_t$ such that
\[
\EXP\left[   \sum_{t=1}^T \sum_{j\in [K]} q_{t,j} X_{t,j} +
    \sum_{t=1}^T  h\left(\bq_t\right) \right] \ge \max_{\bN} T \left(  \left\langle \frac{\bN}{T}, \wh\bmu_T
     \right\rangle + h\left(\frac{\bN}{T}\right) \right) - \gamma_T
 \]
with $\gamma_T= (K+1)\sqrt{2T}\left(1+ \sqrt{\log\frac{\sqrt{2T}}{K+1}}\right)$.
In fact, we prove the slightly stronger bound
\[
 \EXP\left[ \sum_{t=1}^T \sum_{j\in [K]} q_{t,j} X_{t,j} +
    \sum_{t=1}^T  h\left(\bq_t\right) \right] \ge \max_{\bq\in \Delta_K} T \left(  \left\langle \bq, \wh\bmu_T
     \right\rangle + h\left(\bq\right) \right)- \gamma_T~.
\]
In order to construct such a forecaster $\bq_t$, we discretize
$\Delta_K$. Let $\epsilon>0$ and let $\bq^{(1)},\ldots, \bq^{(M)}$ be
a minimal $\epsilon$-cover of $\Delta_K$ with respect to the $\ell_1$
distance. Using the fact that a minimal $\epsilon$-cover is an
$\epsilon/2$ packing and a standard volumetric estimate, we get that
$M \le \epsilon^{-(K-1)}$. Also, by the Lipschitz property of $h$, 
\begin{equation}
\label{eq:discr}
   \max_{i\in [M]} T \left(  \left\langle \bq^{(i)}, \wh\bmu_T
     \right\rangle + h\left(\bq^{(i)}\right) \right)
   \ge
   \max_{\bq\in \Delta_K} T \left(  \left\langle \bq, \wh\bmu_T
     \right\rangle + h\left(\bq\right) \right)  - 2T\epsilon~.
\end{equation}
We treat $\bq^{(1)},\ldots, \bq^{(M)}$ as \emph{experts} and use the
exponentially weighted average forecaster. In order to do this,
we use the estimates
\[
  \wh{X}_{t,j}= 1 - \frac{1 - X_{t,j}}{q_{t,j}} \one_{J_t=j}~, \quad
  j \in [K]~,
\]
where $q_{t,j}$ is the probability our forecaster plays arm $j$ in round $t$.
Observe that $\wh{X}_{t,j}$ may be computed for all $j$ since
$X_{t,J_t}$ is observed after selecting arm $J_t$ at time $t$. Also
note that $\EXP_{J_t \sim \bq_t} \wh{X}_{t,j} = X_{t,j}$ and  $1-\wh{X}_{t,j} \ge 0$
To each expert $\bq^{(i)}$ we assign
the ``loss estimate''
\[
 \wh\ell_t(\bq^{(i)})=   \sum_{j\in [K]} q^{(i)}_j (1-\wh{X}_{t,j}) -
 h(\bq^{(i)})
 =   1 - \sum_{j\in [K]} q^{(i)}_j \wh{X}_{t,j} - h(\bq^{(i)})~.
\]
Now we are prepared to define the proposed forecaster.
At each time instance $t=1,\ldots,T$, $\bq_t$ is chosen as a weighted average
of the experts $\bq^{(1)},\ldots, \bq^{(M)}$, weighted by the
distribution $p_{t,1},\ldots,p_{t,M}$ so that $\bq_t = \sum_{i\in [M]}
  p_{t,i} \bq^{(i)}$, where
for $t=1$, $p_{t,i}=1/M$ for all $i\in [M]$ and for $t >1$,
\[
  p_{t,i} = \frac{w_{t-1,i}}{W_{t-1}}~,
\]
where
\[
  w_{t-1,i} =     \exp\left(-\eta \sum_{s=1}^{t-1}
       \wh\ell_s(\bq^{(i)})\right)  \quad \text{and} \quad W_{t-1} =
     \sum_{i\in [M]} w_{t-1,i}~.
   \]
Here $\eta>0$ is a tuning paramenter.
We proceed by following analysis of the exponentially weighted average
forecaster (see, e.g., \cite{CeLu06}).
On the one hand, 
\begin{eqnarray*}
  \log \frac{W_T}{W_0} & = &
           \log \left( \sum_{i\in [M]}  \exp\left(-\eta \sum_{t=1}^T
                             \wh\ell_t(\bq^{(i)})\right)\right) - \log M    \\
                       & \ge &
                           -\eta \sum_{t=1}^T\wh\ell_t(\bq^{(i)})
                               - \log M
\end{eqnarray*}
for all $i\in [M]$.

On the other hand,
\begin{eqnarray*}
  \log \frac{W_T}{W_0} & = &
                             \sum_{t=1}^T\log \frac{W_t}{W_{t-1}}  \\
  & = &
                             \sum_{t=1}^T \log \left( \sum_{i\in [M]} p_{t,i}
        e^{-\eta \wh\ell_t(\bq^{(i)})} \right)   \\
                       & \le &
          \sum_{t=1}^T \log \left( 1 - \eta \sum_{i\in [M]} p_{t,i}
                               \wh\ell_t(\bq^{(i)})+ \eta^2 \sum_{i\in [M]} p_{t,i}
                               \wh\ell_t(\bq^{(i)})^2 \right)  \\
 & & \text{(using $e^{-x} \le 1-x+x^2$ for $x\ge -1$ and 
     $\wh\ell_t(\bq^{(i)})\ge -1$ by construction)} \\
                       & \le &
   - \eta   \sum_{t=1}^T \sum_{i\in [M]} p_{t,i}
                               \wh\ell_t(\bq^{(i)}) + \eta^2
                               \sum_{t=1}^T \sum_{i\in [M]} p_{t,i}
                               \wh\ell_t(\bq^{(i)})^2~.
\end{eqnarray*}
Comparing the upper and lower bounds for $\log(W_T/W_0)$, 
we obtain that, for every $i \in [M]$
\begin{equation}
\sum_{t=1}^T \sum_{i\in [M]} p_{t,i}
\wh\ell_t(\bq^{(i)})   \le 
\sum_{t=1}^T\wh\ell_t(\bq^{(i)})
                               + \frac{\log M}{\eta} + \eta
                               \sum_{t=1}^T \sum_{i\in [M]} p_{t,i}
                               \wh\ell_t(\bq^{(i)})^2~. \label{eqn:decomp}
\end{equation}
Then, observe that $\EXP_{J_t\sim \bq_t} \wh \ell_t(\bq^{(i)}) =   1 - \sum_{j\in
  [K]} q^{(i)}_j X_{t,j} - h(\bq^{(i)})$, $\EXP_{J_t\sim \bq_t} \sum_{i\in [M]} p_{t,i} \wh\ell_t(\bq^{(i)}) = 1- \sum_{j\in
  [K]} q_{t,j} X_{t,j} - \sum_{i \in [M]} p_{t,i} h(\bq^{(i)})$ and for each $t\in [T]$,
\begin{eqnarray*}
\EXP_{J_t\sim \bq_t} \sum_{i\in [M]} p_{t,i}
                               \wh\ell_t(\bq^{(i)})^2
  & = &
        \sum_{i\in [M]} p_{t,i}  \EXP_{J_t\sim \bq_t}  \left(
        \sum_{j\in [K]} q^{(i)}_j (1 -\wh{X}_{t,j}) - h(\bq^{(i)})\right)^2
  \\
  & \le & 2 + 2 \sum_{i\in [M]} p_{t,i}  \EXP_{J_t\sim \bq_t}  \left(
          \sum_{j\in [K]} q^{(i)}_j (1-\wh{X}_{t,j}) \right)^2  \\
  & = &
        2 + 2 \sum_{i\in [M]} p_{t,i}  \EXP_{J_t\sim \bq_t}  \left(
          \sum_{j\in [K]} q^{(i)}_j \frac{1-X_{t,j}}{q_{t,j}} \one_{J_t=j} \right)^2  \\
  & = &
        2 + 2 \sum_{i\in [M]} p_{t,i}   \sum_{j\in [K]} q_{t,j} \left(
        q^{(i)}_j \frac{1-X_{t,j}}{q_{t,j}} \right)^2  \\
  & \le & 
        2 + 2 \sum_{i\in [M]} p_{t,i}   \sum_{j\in [K]} 
          \frac{q^{(i)}_j}{q_{t,j}}   \\
 & = & 2(K+1) \qquad \text{(using  $\bq_t=\sum_{i\in   [M]} p_{t,i}\bq^{(i)}$)}~.
\end{eqnarray*}
So we can rearrange the expression in \eqref{eqn:decomp} to get
\begin{eqnarray*}
\lefteqn{
     \sum_{t=1}^T \sum_{j\in [K]} q_{t,j} X_{t,j}+ \sum_{t=1}^T
  \sum_{i\in [M]} p_{t,i} h(\bq^{(i)})    } \\
  &  \ge &
    \max_{i \in [M]}  \left(       \sum_{t=1}^T \sum_{j\in
       [K]} q^{(i)}_j X_{t,j} + T h(\bq^{(i)})  \right) - \frac{\log
           M}{\eta} - 2(K+1)\eta T  \\
  & \ge &
   \max_{\bq\in \Delta_K} T \left(  \left\langle \bq, \wh\bmu_T
     \right\rangle + h\left(\bq\right) \right)  - 2T\epsilon - \frac{(K-1)\log\frac{1}{\epsilon}}{\eta} - 2(K+1)\eta T
  \\
  & & \text{(by \eqref{eq:discr}) and since $M\le \epsilon^{-(K-1)}$)}
  \\
  & = &
 \max_{\bq\in \Delta_K} T \left(  \left\langle \bq, \wh\bmu_T
     \right\rangle + h\left(\bq\right) \right) - 2T\epsilon -
        (K+1)\sqrt{2T\log\frac{1}{\epsilon}}  
  \\
  & & \text{(choosing $\eta = \sqrt{\log(1/\epsilon)/(2T)}$)}  \\
  & = &
        \max_{\bq\in \Delta_K} T \left(  \left\langle \bq, \wh\bmu_T
     \right\rangle + h\left(\bq\right) \right) -
        (K+1)\sqrt{2T}\left(1+ \sqrt{\log\frac{\sqrt{2T}}{K+1}}\right)
\end{eqnarray*}
with the choice $\epsilon = (K+1)/\sqrt{2T}$.
The proof is finished by noting that, by concavity of $h$,
\[
  \sum_{i\in [M]} p_{t,i} h(\bq^{(i)}) \le h\left(\sum_{i\in [M]}
    p_{t,i}\bq^{(i)}\right) = h(\bq_t)~,
\]
and therefore
\[
     \sum_{t=1}^T \sum_{j\in [K]} q_{t,j} X_{t,j}+ \sum_{t=1}^T
     h(\bq_t)
     \ge  \max_{\bq\in \Delta_K} T \left(  \left\langle \bq, \wh\bmu_T
     \right\rangle + h\left(\bq\right) \right) -
        (K+1)\sqrt{2T}\left(1+ \sqrt{\log\frac{\sqrt{2T}}{K+1}}\right)~.
      \]
Combining this with \eqref{eq:azuma} implies the announced regret
bound.
\end{proof}

\begin{remark}
Since the set of experts needs to cover the simplex, the proposed
algorithm may not be computationally efficient. A possible way to
construct computationally efficient low-regret forecasters
is to exploit the concavity of $h$ and use ideas from bandit
convex optimization (e.g., \citet{bubeck2017kernel}).
However, it is unclear whether such an algorithm achieves the same
dependence on $K$. This is left for future research.
\end{remark}

\section{Loyalty points model: coupon rewards }
\label{sec:loycoup}

In this section we study the loyalty points model with ``coupons'' fidelity rewards.
Recall that in this model, associated to each arm $j\in [K]$, are a positive integer $\rho_j$ (the period length)
and a real number $r_j \in [0,1]$ and the player receives the extra fidelity reward $r_j$ after each
(not necessarily consecutive) $\rho_j$ plays of arm $j$.  

\subsection{Minimal sufficient sets}

Once again, in order to understand regret, the first step is to determine the class of minimal sufficient sets.
To this end, we need to
understand the set of types $\bN$ that are maximizers of
the function $\sigma(\bmu,\bN)$ for some $\bmu \in [0,1]^K$.
This is not a simple task because of the nonlinear nature of the coupon rewards.
Note that
\[
   \sigma(\bmu,\bN)= \sum_{j=1}^k \left( \mu_j N_j + r_j \left\lfloor \frac{N_j}{\rho_j}\right\rfloor\right)~.
\]
The presence of the integer part in the terms $\left\lfloor N_j/\rho_j \right\rfloor$ means that the set
of possible optimizers may be quite rich and complex. However, in some cases, the class of
minimal sufficient sets is simple. For example, consider the simplest case when the period
lengths $\rho_j$ associated to all arms $j\in [K]$ are the same, say equal to $\rho$.
If $T$ happens to be divisible by $\rho$, then it is easy to see that playing an arm $j$ that maximizes
$\mu_j + r_j/\rho$ forever maximizes the reward. (In other words, $N_j=T$ for a maximizing arm $j$
and $N_i=0$ for all other arms is an optimal type.) Hence, in this case, the unique minimal
sufficient set of sequences is the set $\J_0$ of single-arm
strategies. 
However, even if $T$ is not an integer multiple of the
period length $\rho$, if $\rho_j=\rho$ for all $j\in [K]$, then for some $\bmu \in [0,1]^K$ the optimal type is such that
an arm maximizing $\mu_j + r_j/\rho$ is played $\rho \lfloor T/\rho \rfloor$ times
and a (possibly different) arm maximizing $\mu_j$ is played the remaining
$T-\rho \lfloor T/\rho \rfloor$ times. Thus, in this case, the class of minimal sufficient sets consists of
all sequences that are either single-arm strategies or such that only two arms are played, one of them
$\rho \lfloor T/\rho \rfloor$ times and the other $T-\rho \lfloor T/\rho \rfloor$ times.
When the period lengths are different, the optimizing sequences may have a more complicated
structure.\footnote{Note that the optimization problem is closely related to (though not completely equivalent to)
a knapsack problem where each item
$j$ has size $\rho_j$ and reward $\rho_j\mu_j + r_j$ and the total capacity of the knapsack is $T$.}

Nevertheless, the above argument easily generalizes to possibly different period lengths as stated in the next
lemma. 

\begin{lemma}
\label{lem:couponminsuffsets}
Let $\ol{\rho}$ denote the least common multiple of the period lengths $\rho_1,\ldots,\rho_K$.
If $T$ is an integer multiple of $\ol{\rho}$, then the unique minimal
sufficient set of sequences is the set $\J_0$ of single-arm
strategies.
More generally, every minimal sufficient set $\J\in\C_f$ is such that every sequence
in $\J$ has type $\bN=(N_1,\ldots,N_K)$ such that for some $j\in [K]$, $N_j\ge \ol{\rho}\lfloor T/\ol{\rho}\rfloor$.
\end{lemma}

In other words, even though single-arm strategies are not necessarily optimal,
every $\bj\in \bigcup_{\J\in\C_f}\J$ is ``almost'' a single-arm strategy in the sense that
it differs from one in at most $T-\ol{\rho}\lfloor T/\ol{\rho}\rfloor \le \ol{\rho}$ positions.
This fact will help us design strategies with small strong regret in the adversarial version
of the problem.

\subsection{Stochastic rewards}
\label{sec:stoccoup}

The observation made in Lemma \ref{lem:couponminsuffsets} suggests a simple way of achieving small
regret in the stochastic loyalty-points bandit problem with coupon rewards. One may simply
aim at  accumulating a reward that is not much smaller than
$\max_{j\in [K]} \left( \mu_j + \frac{r_j}{\rho_j} \right)$. This may be achieved by applying
a regret-minimization strategy using the augmented rewards
$\wh{Y}_{t,j} = X_{t,j} + \frac{r_j}{\rho_j}$.
Suppose that one plays according to a standard $K$-armed bandit strategy based on
the rewards $\wh{Y}_{t,j}$, and that the expected cumulative regret of this strategy compared to  $T \max_{j\in [K]} \left( \mu_j + \frac{r_j}{\rho_j} \right)$ is bounded by a quantity $\epsilon_T$.
This means that if $\wh{N}_1,\ldots,\wh{N}_K$ denote the (random) number of times each of
the $K$ arms is pulled up to time $T$, then
\[
  \sum_{j=1}^K \EXP \wh{N}_j\left( \mu_j + \frac{r_j}{\rho_j} \right) \ge
     T \max_{j\in [K]} \left( \mu_j + \frac{r_j}{\rho_j} \right) - \epsilon_T~.
\]
This implies that the expected fidelity-augmented reward of the strategy  satisfies
\begin{eqnarray*}
 \sum_{j=1}^k \EXP  \left( \mu_j \wh{N}_j + r_j \left\lfloor \frac{ \wh{N}_j}{\rho_j}\right\rfloor\right)
  & \ge &
          \sum_{j=1}^k \EXP  \left( \mu_j \wh{N}_j + r_j \left(\frac{ \wh{N}_j}{\rho_j} - 1\right) \right) \\
  & \ge & 
          T \max_{j\in [K]} \left( \mu_j + \frac{r_j}{\rho_j} \right) - \epsilon_T - K   \\
  & \ge & \max_{\bN} \sum_{i=1}^k N_i \left( \mu_i + \frac{r_i}{\rho_i} \right) - \epsilon_T - K   \\
    & \ge & \max_{\bN} \sigma(\bmu,\bN) - \epsilon_T - K~.
\end{eqnarray*}
In other words, the regret of the strategy is at most $\epsilon_T+K$.
For example, by using a standard UCB algorithm \citep{LaSz20}, one may take
\[
\epsilon_T = \sum_{j\in [K]: \Delta_j>0} \frac{ 16\log(T)}{ \Delta_j}+ 3K~,
\]
where
\[
  \Delta_j= \max_{i\in [K]} \left(\mu_i + \frac{r_i}{\rho_i}\right) - \left( \mu_j + \frac{r_j}{\rho_j} \right)
\]
are the ``gap'' parameters, see \cite{LaSz20}. Summarizing, we have the following:

\begin{theorem}
\label{thm:regstolpcoup}
Consider the stochastic loyalty-points bandit problem with coupon rewards.
There exists a strategy of play whose expected regret satisfies 
\[
  \E[ \Reg_T] \le \sum_{j\in [K]: \Delta_j>0} \frac{ 16\log(T)}{ \Delta_j}+ 4K~.
\]
Consequently, the worst case regret of this strategy is $ O(\sqrt{KT\log(T)})$.
\end{theorem}

\subsection{Adversarial rewards}
\label{sec:advcoup}

Equipped with our understanding of the class of minimal sufficient sets and having derived a small-regret
strategy for the stochastic version of the problem, it is now easy to design strategies that work well
in the adversarial case. The main take-home message from the previous sections is that
(a) single-arm strategies are almost optimal in the presence of coupon rewards; (b) playing a standard
bandit strategy based on the augmented rewards $X_{t,j}+ r_j/\rho_j$ is nearly optimal.
We show here that the same principles work in the adversarial case as well.

\begin{theorem}
\label{thm:regadlpcoup}
Consider the adversarial loyalty-points bandit problem with coupon rewards.
There exists a randomized policy $\pi$ whose expected weak, mean and strong regret satisfy
\[
   \EXP \Reg_T^{\flat}(\pi) \le  \EXP \Reg_T^{\natural}(\pi) \le 4\sqrt{TK\log K} + K~,
\]
and
\[
  \EXP \Reg_T^{\sharp}(\pi) \le 4\sqrt{TK\log K}+ \ol{\rho}+ K~.
\]
\end{theorem}

\begin{proof}
  The argument for bounding the mean regret is similar to that of Theorem \ref{thm:regstolpcoup}.
  Suppose that one plays a standard (randomized) adversarial bandit strategy based on the augmented rewards
  $\wh{Y}_{t,j} = X_{t,j} + \frac{r_j}{\rho_j}$. Denote by $J_t$ the arm selected by the strategy at time $t$
  and by $\wh{N}_j=\sum_{t=1}^T \IND_{J_t=j}$ the number of times the strategy plays  arm $j$ up to time $T$.
  Recall also that     $\wh\mu_{T,j} = \frac{1}{T} \sum_{t=1}^T X_{j,t}$ denotes the empirical average of the
  base rewards of arm $j$.
  Assume that the strategy comes with the regret guarantee
\[
      \EXP \sum_{t=1}^T\wh{Y}_{t,J_t}\ge \max_{j\in K} \sum_{t=1}^T \wh{Y}_{t,J_t} - \epsilon_T~.
    \]
    For example, by using the {\sc exp3} algorithm of \cite{auer1995gambling}, one may take $\epsilon_T= 4\sqrt{TK\log K}$
    (see, e.g., \cite[Theorem 11.1]{LaSz20}).
    Then the expected cumulative reward of this strategy equals
\begin{eqnarray*}
  \EXP \sum_{t=1}^TX_{t,J_t} + \sum_{j=1}^K r_j \left\lfloor \frac{\wh{N}_j}{\rho_j} \right\rfloor
  & \ge &
          \EXP \sum_{t=1}^TX_{t,J_t} + \sum_{j=1}^K r_j  \frac{\wh{N}_j}{\rho_j} - K    \\
  & = &
          \EXP \sum_{t=1}^T\wh{Y}_{t,J_t} - K    \\
  & \ge & \max_{j\in K}\sum_{t=1}^T \left( X_{t,j} + \frac{r_j}{\rho_j} \right) - K- \epsilon_T \\
  & = & T \max_{j\in K} \left( \wh{\mu}_{T,j} + \frac{r_j}{\rho_j} \right) - K - \epsilon_T \\
  & = & \max_{\bN} \sum_{i=1}^K N_i \left( \wh{\mu}_{T,i} + \frac{r_i}{\rho_i} \right) - K - \epsilon_T \\
  & = & \max_{\bN} \sigma(\wh\bmu_T,\bN) - K - \epsilon_T~,
\end{eqnarray*}
proving the bound for the mean regret.
To prove the upper bound for the strong regret, notice that by Lemma \ref{lem:couponminsuffsets},
\[
  \max_{\J\in \C_f} \max_{ \bj \in \J} S_t(\bj) \le \max_{ \bj \in \J_0} S_t(\bj) + \ol{\rho}
  = \max_{j \in [K]}   \sum_{t=1}^T \left( X_{t,j} + \left\lfloor \frac{r_j}{\rho_j}  \right\rfloor\right)  + \ol{\rho}
  \le  \max_{j \in [K]}   \sum_{t=1}^T \left( X_{t,j} + \frac{r_j}{\rho_j}  \right) + \ol{\rho}~.
\]
But we have already proved above that
\[
  \max_{j \in [K]}   \sum_{t=1}^T \left( X_{t,j} + \frac{r_j}{\rho_j}  \right) \le
  \EXP \sum_{t=1}^TX_{t,J_t} + \sum_{j=1}^K r_j \left\lfloor \frac{\wh{N}_j}{\rho_j} \right\rfloor + K + \epsilon_T~,
\]
concluding the proof.
\end{proof}

\section{Subscription model: increasing fidelity rewards}

In the rest of the paper we discuss what we call the \emph{subscription model}.
Recall from the introduction that in this model the fidelity rewards depend on the current
number of consecutive plays of the selected arm --- as opposed to the loyalty points models
where the fidelity rewards are a function of the total number of times the selected arm
has been played in the past. Just like in the case of the loyalty points model, here we also consider
three types of fidelity rewards: increasing, decreasing, and ``coupon'' rewards.

We begin the discussion by considering increasing fidelity rewards. More precisely, we assume that the fidelity reward is a nondecreasing function of the number of consecutive plays of an arm prior to the current time step.  Recall the definition $Q_{t,j} = \one_{\{J_{t} = j \}} ( t - \max \{ s \leq t \,:\,J_s = j, J_{s-1} \neq j\})$ as the number of consecutive plays of arm $j$ up to time $t$ and assume that the reward is of the form $Y_{t,j} = X_{t,j} + f_j (Q_{t,j})$, where $X_{t,j}$ is the base reward and $f_j$ is the (known) fidelity function associated with arm $j$ that is assumed to be nondecreasing. We may assume, without loss of generality, that $f_j(0)=0$ for all $j\in [K]$.

\subsection{Minimal sufficient sets}
\label{sec:minsuff_subinc}

It is clear that for the stochastic subscription bandit problem with increasing fidelity rewards, there is always an optimal policy
that is a single-arm policy. Indeed, a single-arm policy
that plays an arm $j^* \in \argmax_{1 \leq j \leq K} \{ \mu_jT + F_j(T)\}$ for all $T$ time steps maximizes total reward.
(Recall that $F_j(t)=\sum_{s=1}^tf_j(t)$ denotes the cumulative fidelity reward of playing arm $j$
during $t$ consecutive steps.)
Hence, the class of minimal and sufficient sets contains the single set $\J_0$ of single-arm strategies.

\subsection{Stochastic rewards}
\label{sec:stocincsub}

We know that in the stochastic bandit problem with increasing fidelity rewards in the subscription model, the reward of the optimal strategy is $\max_{1 \leq j \leq K} \{ \mu_jT + F_j(T)\}$, achieved by playing a single arm $j^*$ for all $T$ rounds.

Since the optimal policy is the same as in the standard stochastic bandits problem, one may imagine that similar learning algorithms can be applied here. However, note that the usual regret-minimizing strategies (such as {\sc ucb}) keep exploring
all through their run. This involves switching to sub-optimal arms in order to
make sure that unlikely events of good arms collecting small rewards get detected. 
Such strategies are not suitable for the subscription model under
increasing fidelity rewards, since a single switch of arms resets the count $Q_{t,j}$ to zero,
resulting in a significant loss of fidelity rewards. In particular, any strategy that
switches an arm after time $\epsilon T$ (for a constant $\epsilon \in (0,1)$) must
suffer a regret of at least $F_{j^*}(\epsilon T)$ which is of order $\Omega(T)$ except
in trivial cases.

However, an expected reward that is not much smaller than that of the optimal strategy may be achieved by any strategy that is able to identify an arm achieving the optimum early on.  Indeed, strategies designed for \emph{best-arm identification} may be used directly to achieve sublinear expected regret in the stochastic subscription bandits problem with decreasing fidelity reward. In best-arm identification problems, the goal of the
learner is to learn the identity of the arm with the highest expected reward.
Instead of reviewing the growing literature of best-arm identification, we
show how such a simple strategy may be adopted, in a straightforward manner, to obtain strategies
with sublinear regret.
We refer the reader to the book Lattimore and Szepesv{\'a}ri \cite{LaSz20} for an exhaustive survey.

The notion of \emph{simple regret} proves to be useful in our approach.
To define the simple regret, we first define the sub-optimality gap of arm $j$ as $\wt \Delta_j = \mu_{j ^*} + F_{j^*}(T)/T - (\mu_j + F_j(T)/T)$ where $j^* = \argmax_{1 \leq j \leq K} (\mu_j + F_j(T)/T)$. (Note that in the case where the fidelity functions are the same across all arms, $\wt \Delta_j = \mu_{j^*} - \mu_j = \Delta_j$.) The \emph{simple regret} of the best-arm identification strategy outputting arm $J$ is defined as
\[
    \EXP \wt \Delta_J~.
\]

Suppose that one plays a best-arm identification strategy during the first $t_0 < T$
rounds, besed on the fidelity-augmented rewards $X_{t,j}+F_j(T)/T$,
at the end of which the algorithm identifies an arm $J \in [K]$.
For the rest of the rounds $t\in \{t_0+1,\ldots,T\}$, the strategy plays arm $J$,
independently of the rewards. During these $T-t_0$ steps, the strategy accumulates
a total expected reward of $\EXP [(T-t_0) \mu_J + F_J(T-t_0)]$.

The expected regret of the above-defined strategy may be bounded in terms of the
simple regret, since
\begin{eqnarray*}
  \E[ \Reg_T] & \le & T \mu_{j ^*} + F_{j^*}(T) - \EXP \left[  (T-t_0) \mu_J + F_J(T-t_0) \right] \\
              & \le & T \mu_{j ^*} + F_{j^*}(T) - \EXP \left[  T \mu_J + F_J(T) \right]  + 2t_0 \\
  & = &   2T\EXP \wt \Delta_J  +2t_0~.
\end{eqnarray*}
The simplest best-arm identification strategy samples all $K$ arms $\lfloor t_0/K\rfloor$
times and outputs an arm that maximizes the empirical reward accumulated over this time.
This simple strategy achieves an expected simple regret bounded by $C\sqrt{(k\log k)/t_0}$
for a universal constant $C$ (see \cite{LaSz20}). Substituting this into the upper bound above and
choosing $t_0 \sim T^{2/3}(K\log K)^{1/3}$, we obtain the following.

\begin{theorem}
\label{thm:regstosubinc}
Consider the stochastic bandit problem in the subscription model with increasing fidelity rewards.
There exists a strategy of play whose expected regret satisfies 
\[
  \E[ \Reg_T] \le C T^{2/3}(K\log K)^{1/3}~,
\]
where $C$ is a numerical constant.
\end{theorem}

Of course, the naive algorithm used here is not optimal and more sophisticated best-arm
strategies may be used to improve the regret bound. In particular, one may get a bound that is
logarithmic in $T$ but depends on the sub-optimality gaps $\wt \Delta_j$.
However, the use of such improved strategies involves the same key idea and
we prefer to present the idea in its simplest form.

\subsection{Adversarial rewards}
\label{sec:advincsub}

Next we consider the adversarial version of the problem. As noted in
Section \ref{sec:minsuff_subinc}, the only class of
minimal and sufficient sets contains the single set $\J_0$ of single-arm strategies.
This implies that the notions of weak and strong regrets coincide as the
learner simply competes with the best single-arm strategy.

It is easy to see that one cannot expect to achieve sublinear regret in general.
In fact, for any  sequence $\bj=(j_1,\ldots,j_T)$ of actions, the corresponding
cumulative reward $S_T(\bj)$ is at most the analogous cumulative reward in
the loyalty-points model. Hence, minimizing regret with respect to
the class of single-arm strategies is at least as hard in the subscription model as in
the loyalty-points model. In particular, Theorem \ref{thm:TregLoy} immediately
implies the following:

\begin{theorem} 
\label{thm:TregSub}
Consider the adversarial bandits problem in the subscription model with increasing fidelity rewards, where the fidelity
function is the same for every arm. Define $\delta = f(7T/8)-f(T/8)$.
Then for every policy $\pi$ there exists a sequence of rewards such that the regret
satisfies
\[ 
 \Reg_T(\pi) \ge \frac{T\delta}{40}~. 
\]
\end{theorem}

The theorem shows that if the fidelity function grows substantially between
(say) $T/8$ and $7T/8$, one cannot expect sublinear regret. If this is not the case,
sublinear regret may be possible, depending on the fidelity function. 
In the rest of the section we take a closer look at the special but important case
when the fidelity reward follows a step function.
More precisely, consider the case where the fidelity functions
take the form
\[ f_j(t) = \left \{ \begin{array}{ll}
					1 & \quad \text{ if } t \geq m \\
					0 & \quad \text{ if } t >m
                     \end{array} \right. \]
for all $j\in [K]$ and for some positive integer $m \le T$.
We refer to $f_j(t)$ as an $m$-step function.

Theorem \ref{thm:TregSub} shows that when $m \ge T/8$  (or more generally, when
$m =\Omega(T)$), sublinear regret cannot be guaranteed. However, for small values of $m$,
small regret is achievable.

To see why, consider first the extremal case when $m=1$. 
In this case, the only time when the player does not receive a fidelity reward
is when the player switches actions. 
Hence, the problem is equivalent to a bandit problem with switching costs
that has been studied in the adversarial setting by
Dekel, Ding, Koren, and Peres \cite{dekel2013bandits}
and Cesa-Bianchi, Dekel, and Shamir \cite{cesa2013online}.
In particular, Theorem 1 of \cite{dekel2013bandits}, shows that for every randomized strategy in the bandits with switching costs problem, there exists a sequence of rewards such that the regret is $\Omega(T^{2/3}K^{1/3})$ for $K\leq T$.

This lower bound is, in fact, achievable. In fact, an easy ``batch'' modification of the EXP3
algorithm shows that regret of the order of $O(T^{2/3} K^{1/3}m^{1/3})$ is achievable
when the fidelity reward is an $m$-step function.
This is stated in the next theorem. %

\begin{theorem} \label{thm:Bexp3incsub}
  Consider the adversarial fidelity bandits problem in the subscription model
  with the $m$-step fidelity function. There exists a randomized strategy whose expected
  (weak and strong) regret satisfies
\[
  \E[\Reg_T] = O((K\log K)^{1/3} T^{2/3} m^{1/3})~.
\]
\end{theorem}

In particular, sublinear regret
is achievable whenever $m=o(T)$, hence characterizing the range of values of $m$
for which sublinear regret is achievable.

\section{Subscription model: decreasing fidelity rewards}

We continue the study of the subscription model in the case when the
fidelity rewards $f_j$ are \emph{nonincreasing} functions of  $Q_{t,j} = \one_{\{J_{t} = j \}} ( t - \max \{ s \leq t \,:\,J_s = j, J_{s-1} \neq j\})$, the number of consecutive plays of arm $j$ prior to the current time step.

This case is substantially different from the case of increasing fidelity rewards
studied in the previous section. As always, we start by determining the class
of minimal and sufficient sets that allows us to design regret minimization strategies
in the stochastic model and to define regret in the adversarial case.

\subsection{Minimal sufficient sets}
\label{sec:minsuffdecsub}

In order to define the minimal sufficient sets for given nonincreasing
fidelity functions $f_1,\ldots,f_K$, we need to characterize the sequences $\bj \in [K]^T$
that can have a maximal expected reward in the stochastic bandit problem for some values of the expected rewards $\mu_1, \dots, \mu_K$ of the $K$ arms. 
To this end, note that for fixed $\mu_1, \dots, \mu_K$,
it is optimal to play the arm $j^*$ with the largest expected initial reward, that is, $j^*= \argmax_{1 \leq j \leq K} \{\mu_j+ f_j(0)\}$, repeatedly until time $m \in [1,T]$ when $\mu_{j^*} + f_{j^*}(m+1) < \max_{j \neq j^*} \{\mu_j + f_j(0)\}$, that is until its expected reward drops under the initial expected reward of the second best arm. An arm $j^{*(2)} \in  \argmax_{j \neq j^*} \mu_j$ is then played once to allow for the $Q$ value of the optimal arm to reset and then arm $j^*$ is played again for $m$ steps. This is then repeated in every period of $m+1$ steps.

Each such sequence $\bj$ is indexed by a triplet
$(i,k,m) \in [K]\times [K]\times [T-1]$, 
corresponding to periodically playing arm $i$ $m\in \N$ times before arm $k$ is played once
and then repeating it. Denote this sequence by
\begin{equation}
\label{eq:jikm}  
     \bj(i,k,m)= (\underbrace{i,\ldots,i}_\text{$m$ times},k, \underbrace{i,\ldots,i}_\text{$m$ times},k, \cdots )
\end{equation}
Moreover, if the fidelity functions $f_j$ are strictly decreasing, then
every such sequence is the only optimal sequence for some values
of $\mu_1,\ldots, \mu_K$. Hence, in such cases, there is only one minimal sufficient set,
containing all sequences of the form $\bj(i,k,m)$, that is,
\[
   \J= \left\{ \bj(i,k,m): (i,k,m) \in [K]\times [K]\times [T-1]\right\}~. 
\]
Moreover, in all cases, the minimal sufficient set is unique and it is a subset of $\J$ defined above.

\subsection{Stochastic rewards}
\label{sec:stocdecsub}

In this section we show that one may achieve sublinear regret in the stochastic
version of the problem. As it is shown in the previous section, the goal of
the learner is to achieve an expected reward that is not much smaller than
the expected reward corresponding to playing the best sequence of the type $\bj(i,k,m)$,
that is,
\begin{equation}
\label{eq:triple}
  \max_{(i,k,m) \in [K]\times [K]\times [T-1]}  W_T(\mu_i,\mu_k,m)
\end{equation}
where
\begin{eqnarray}
\label{eq:wdef}
 W_T(\mu_i,\mu_k,m)& \defeq &
  \left( T - \left\lfloor \frac{T}{m+1} \right\rfloor\right)\mu_i + \left\lfloor \frac{T}{m+1} \right\rfloor\mu_K   \nonumber \\
& &  +  \left\lfloor \frac{T}{m+1} \right\rfloor \left( F_i(m) + F_k(1) \right)
     + F_i \left( T - (m+1)\left\lfloor \frac{T}{m+1} \right\rfloor\right)~
\end{eqnarray}
is the expected reward of playing arm $i$ $m$ times, then arm $j$ once and repeating this until horizon $T$ in a subscription bandits problem where the expected base rewards are given by $\mu_1,\dots, \mu_K$.
We denote by $(i^*,k^*,m^*)$ a triple achieving the maximum in \eqref{eq:triple}.

The proposed strategy is similar to the one of Section \ref{sec:stocincsub} in that
the first $t_0$ rounds of the game are used purely for exploration. After time $t_0$,
the strategy commits to a triplet $(i,k,m)$ and plays accordingly.

More precisely, up to time $t_0$, the strategy samples each arm $\lfloor t_0/K\rfloor$ times
and computes the empirical average of the observed base reward of each arm. Denote
these estimates by $\wh\mu_1,\ldots,\wh\mu_K$. By a simple use of Hoeffding's inequality,
we have
\begin{equation}
\label{eq:hoeffdingbound}
  \EXP \max_{j\in [K]} |\wh\mu_j - \mu_j |\le \sqrt{\frac{K\log K}{2t_0}}~.   
\end{equation}
Let $\hat{i}$ and $\hat{k}$ denote the arms corresponding to the two highest values
of $\wh\mu_i+f_i(0)$ such that $\wh\mu_{\hat{i}} + f_{\hat{i}}(0) \ge \wh\mu_{\hat{k}}+ f_{\hat{k}}(0)$.
(In case of ties, we may choose arbitrarily.) Moreover, let $\wh{m}$ denote the
corresponding period length, that is,
\[
\wh{m} = \min\left\{m:  \wh\mu_{\hat{i}} + f_{\hat{i}} (m+1) <  \wh\mu_{\hat{k}} + f_{\hat{k}} (0)  \right\}~.
\]
Equivalently,
\begin{equation}
\label{eq:defofm}  
  (\hat i, \hat k, \hat m) = \argmax_{i,j,m} W_T(\wh\mu_i,\wh\mu_k, m)~.
\end{equation}
Starting from time $t_0+1$, until time $T$,
the strategy plays according to the sequence $\bj(\hat{i},\hat{k},\wh{m})$.

\begin{theorem} \label{thm:regstocsubdec}
  Consider the stochastic bandit problem in the subscription model with decreasing
  fidelity rewards. 
  The expected regret of the explore-then-commit strategy defined above with $t_0=(2T)^{2/3}(K\log K)^{1/3}$
  is bounded by
  \[
     \EXP [ \Reg_T] \le 3 T^{2/3}(K\log K)^{1/3}~.
   \]
\end{theorem}

\begin{proof}
  Conditionally on the rewards observed during the first $t_0$ periods of play,
  the expected reward of the proposed strategy for the remaining $T-t_0$ rounds equals
  \begin{eqnarray*}
 \lefteqn{   
  W_{T-t_0}(\mu_{\hat{i}},\mu_{\hat{k}},\wh{m})    } \\
  & \ge  &
           W_{T}(\mu_{\hat{i}},\mu_{\hat{k}},\wh{m}) - 2t_0  \quad \text{(since all rewards are bounded by $2$)} \\
 & \ge & W_{T}(\wh\mu_{\hat{i}},\wh\mu_{\hat{k}},\wh{m}) - 2t_0 - T \max_{j\in [K]} |\wh\mu_j - \mu_j |\\
 & \ge & W_{T}(\wh\mu_{i^*},\wh\mu_{k^*},m^*) - 2t_0 - T \max_{j\in [K]} |\wh\mu_j - \mu_j |
 \quad \text{(by \eqref{eq:defofm})}  \\
 & \ge & W_{T}(\mu_{i^*},\mu_{k^*},m^*)- 2t_0 - 2  T \max_{j\in [K]} |\wh\mu_j - \mu_j |~.
\end{eqnarray*}
Taking expected values and using \eqref{eq:hoeffdingbound}, we get that
\[
  \EXP [\Reg_T] \le W_{T}(\mu_{i^*},\mu_{k^*},m^*) - \EXP  W_{T-t_0}(\mu_{\hat{i}},\mu_{\hat{k}},\wh{m})
  \le 2t_0 + 2 T \sqrt{\frac{K\log K}{2t_0}}~.   
\]  
Choosing $t_0 = T^{2/3}(K\log K)^{1/3}/2$, we obtain the stated bound.  
\end{proof}

We remark that, just like in the case of increasing fidelity rewards, one may also
design strategies that yield regret bounds that have a logarithmic dependence on the time
horizon $T$. For example, one may use successive elimination strategies in the
exploration phase to identify the two arms $i^*$ and $k^*$ played by the optimal strategy.
The price to pay is that the regret bound depends on the means $\mu_j$ of the arms. In
particular, they depend on the gap between the expected
reward of the best and second best arm, as well
as on the gap between the expected reward of the second and the third best arms.
In order to keep the discussion simple, we omit these regret bounds from this paper.

\subsection{Adversarial rewards}
\label{sec:advdecsub}

In this section we show that in the adversarial bandit problem in the
subscription model with decreasing fidelity rewards, 
it is possible to achieve sublinear regret. Recall from Section
\ref{sec:minsuffdecsub} that in this model, there is only one minimal
sufficient set, and in all cases it is a subset of the class $\J$
containing all sequences of the type $\bj(i,k,m)$ for $(i,k,m)\in
[K]\times [K] \times [T-1]$. Hence, weak and
strong regret of a policy $\pi$ coincide and, recalling
\eqref{eq:regret}, are bounded as
\[
\Reg_T(\pi) \le \max_{(i,k,m)\in [K]\times [K] \times [T-1]} S_T(\bj(i,k,m))- \wt S_T(\pi)~.
\]
In other words, our goal is to design a strategy whose expected
cumulative reward is close to $\max_{(i,k,m)} S_T(\bj(i,k,m))$. 

Our approach is to treat each triple $(i,k,m)$ as an ``expert'' that
plays according to the sequence $\bj(i,k,m)$ and
design a strategy that competes with the best expert.
To this end, we use a variant of the exponentially weighted average
forecaster that assigns a weight to each expert, based on an
exponential function of its
estimated cumulative reward, and selects a random expert with
probability proportional to its weight.
A difficulty in the subscription bandits problem is that the reward of the strategy depends not only on
the reward of the chosen expert but also on the number of consecutive
plays of the selected arm.
Therefore the reward an expert would gain from playing an arm in a given round may differ from the actual reward the learner receives.
In order to keep such deviations under
control, we draw inspiration from the \emph{lazy label efficient
  exponentially weighted average forecaster} proposed by Cesa-Bianchi,
Lugosi, and Stoltz \cite{CeLuSt04}. The forecaster, at each time
instance, flips a biased coin that comes up heads with probability
$\epsilon$, where $\epsilon$ is a small, appropriately chosen value.
When the coin flip results in heads, the forecaster selects a random
arm, plays it, and updates the estimated cumulative reward of
each expert.  Then the algorithm selects a random expert according to the
exponentially weighted average distribution over the experts, and plays
according to the sequence corresponding to the selected expert
until the next time when the coin flip comes up
heads. This ``lazy'' play guarantees that the strategy switches
between experts at most about $\epsilon T$ times, making the possibly
adverse effect of switching between experts negligible.

More precisely, at every time $t \in \{2,\ldots,T\}$, an independent Bernoulli
random variable $Z_t$ with
        $\PROB\{Z_t=1\} = 1- \PROB\{Z_t=0\}= \epsilon$
 is drawn. If $Z_t=1$, then the strategy simply
 explores by selecting an arm $J_t$ chosen uniformly at random.
(We define $Z_1=1$ to ensure that the strategy explores in the first step.)
 These
 time instances are used to estimate the cumulative reward
 of each expert. In particular, for an expert $(i,k,m)\in [K]\times
 [K]\times [T-1]$, let $j_t(i,k,m) \in [K]$ denote the arm played by
 the expert at time $t$, that is, the $t$-th element of the sequence
 $\bj(i,k,m)$ as defined in \eqref{eq:jikm}. The cumulative 
 reward of the expert, at time $t$, is
\[
  S_t(\bj(i,k,m))=   \sum_{s=1}^t \left(
    X_{s, j_s(i,k,m)} + \Phi(i,k,m)\right)~,
\]
where we define
\[
  \Phi(i,k,m)= \frac{1}{T}
  \left(
  \left\lfloor \frac{T}{m+1} \right\rfloor \left( F_i(m) + F_k(1) \right)
     + F_i \left( T - (m+1)\left\lfloor \frac{T}{m+1} \right\rfloor\right)  \right)
 \]
as the normalized total fidelity reward received by the expert.

$S_t(\bj(i,k,m))$ is estimated by
\[
\wh{S}_t(i,k,m) = \sum_{s=1}^t \wh{Y}_{s, j_s(i,k,m)}
\]
where
\[
  \wh{Y}_{s, j_s(i,k,m)}\defeq 2 - \frac{K}{\epsilon} \IND_{Z_s=1} \IND_{J_s=j_s(i,k,m)}  (2-X_{s, j_s(i,k,m)}-\Phi(i,k,m))
\]
is the estimated reward of expert $(i,k,m)$ at time $s$, where $J_s$ indicates the arm played by the learner in round $s$.
Note that the factor $K/\epsilon$ guarantees that 
$\EXP \wh{Y}_{s, (i,k,m)} =  X_{s, j_s(i,k,m)}+ \Phi(i,k,m)$ and therefore $\wh{S}_t(i,k,m)$ is an
unbiased estimator of the cumulative total reward of the expert.
It is convenient to introduce the estimated ``losses''
\[
  \wh\ell _{s, (i,k,m)} \defeq 2-\wh{Y}_{s, (i,k,m)} = \frac{K}{\epsilon} \IND_{Z_s=1} \IND_{J_s=j_s(i,k,m)}  (2-X_{s, j_s(i,k,m)}-\Phi(i,k,m))~.
\]
Note that $\wh\ell _{s, (i,k,m)}$ is always nonnegative, a property that is crucial for the proof below. 

In the time instance
immediately following an exploration step -- unless the Benoulli
random variable again equals $1$ --, the learner selects an expert 
$(I_t,K_t,M_t)\in [K]\times [K]\times [T-1]$ at random,
based on the exponentially weighted average distribution.
More precisely,
\[
  \PROB_{t-1}\left\{ (I_t,K_t,M_t) = (i,k,m) \right\} = p_{t,(i,k,m)}
  \defeq \frac{w_{t-1,(i,k,m)}}{W_{t-1}}
\]
where $\PROB_{t-1}$ denotes conditional probability given the past and
\[
  w_{t,(i,k,m)} = \exp(\eta \wh{S}_t(\bj(i,k,m)) )
\]
and
\[
    W_t = \sum_{(i,k,m)\in [K]\times [K]\times [T-1]} \exp(\eta
       \wh{S}_t(\bj(i,k,m)))~,
\]  
Then the learner follows the expert $(I_t,K_t,M_t)$
until the next time the Bernoulli variable takes value $1$. More precisely,
if for some $t$, $Z_t=1$ and $t'=\min\{s>t: Z_s=1\}$, then
for all time instances $s\in \{t,\ldots,t'-1\}$, the learner plays the
arms according to positions $t,\ldots,t'-1$ of the sequence
$\bj(I_t,K_t,M_t)$, as defined in \eqref{eq:jikm}.
The algorithm is described in detail in Figure \ref{alg:advsubdec}.

The next theorem establishes sublinear regret for the performance of the proposed strategy.
In particular, it shows that the expected regret grows at a rate at
most $\wt O(T^{2/3})$,.

 \begin{figure}[t]
\centering
 \fbox{ \parbox{0.85\textwidth}{
 \begin{algorithmic}[1]
 \renewcommand{\algorithmicrequire}{\textbf{Input:}}
 \renewcommand{\algorithmicensure}{\textbf{Initialize:}}
 \algsetup{indent=1.5em}
 \REQUIRE  $T$ (horizon), $\epsilon \in (0,1)$ (sampling probability),
 $\eta >0$ (learning rate).
 \ENSURE   $w_{0,(i,k,m)}=1$. %
\STATE Sample $T$ independent Bernoulli$(\epsilon)$ random variables $Z_1,\dots, Z_T$ .
\STATE Set $Z_1=1$
 \FORALL{ $t=1, \dots, T$}
 \IF{$Z_t=1$,}
 \STATE Select a uniformly random arm $J_t$ and observe $X_{t,J_t}$;
 \STATE Compute $w_{t,(i,k,m)} = \exp(\eta \wh{S}_t(\bj(i,k,m)) )$ 
 $\forall (i,k,m)\in [K]^2 \times [T-1]$;
 \ENDIF
 \IF{$Z_t=0$,}
     \IF{$Z_{t-1}=1$}
 	\STATE Draw a random expert $(I_t,K_t,M_t)$ according to
         the distribution $p_{t,(i,k,m)}= w_{t-1,(i,k,m)}/W_{t-1}$ and play
           according to the $t$-th position of the sequence
           $\bj(I_t,K_t,M_t)$;
           \ELSE
           \STATE Play according to the same expert as at time $t-1$.
         \ENDIF
         \ENDIF
 \ENDFOR
 \end{algorithmic}
 }
 }
 \caption{Lazy strategy for adversarial bandits with decreasing
   fidelity rewards, subscription model}
 \label{alg:advsubdec}
 \end{figure}

\begin{theorem}
\label{thm:advdecsub}
Consider the adversarial fidelity bandits problem in the subscription
model with nonincreasing fidelity rewards. Then, choosing the
parameters of the algorithm as
$\eta = \left(\log(K^2(T-1))/(2TK)\right)^{2/3}$ and $\epsilon = K\sqrt{\eta}$, the expected (weak and strong) regret of the
strategy define above satisfies
\[
   \EXP \Reg_T \le   3(2TK)^{2/3} \log^{1/3}\left(K^2(T-1)\right)~.
 \]
\end{theorem}

\begin{proof}
  Let $\bJ=(J_1,\ldots,J_T) \in [K]^T$ denote the (random) sequence of
  arms played by the strategy.
The strategy is defined such that, except for the times when $Z_t=1$, the
strategy follows the expert $(I_t,K_t,M_t)$, that is, $J_t=\bj_t(I_t,K_t,M_t)$.

  Define the ``pseudo'' reward of the strategy by
\[
     \ol{S}_T\defeq \sum_{t=1}^T \left( X_{t,J_t} + \Phi(I_t,K_t,M_t) \right)
\] 
The key observation is that, since the fidelity rewards are
decreasing, the realized total reward of the strategy is at least
\[
    S_T(\bJ) = \sum_{t=1}^T (X_{t,J_t} +f_{J_t}(Q_{t-1,J_t})) \geq  \ol{S}_T - 2 \sum_{t=1}^T \IND_{Z_t=1}~,
\]  
since $\sum_{t=1}^T \IND_{Z_t=1}$ is the total number of times the
strategy switches experts and the rewards are bounded in $[0,2]$. Hence, it suffices to show that the total
expected pseudo reward $\EXP \ol{S}_T$ is not much smaller than
\[
  \max_{(i,k,m)} S_T(\bj(i,k,m))= \max_{(i,k,m)} \sum_{t=1}^T \left(
    X_{t,j_t(i,k,m)} + \Phi(i,k,m) \right)~.
\]
As it is customary in the analysis of exponentially weighted average
forecasters (see \cite{CeLu06}), we compare upper and lower bounds
for the ``weight'' $W_t$. On the one hand,
\begin{eqnarray*}
    \log \frac{W_T}{W_0} & = & \log \left(\sum_{(i,k,m)}\exp(\eta
                             \wh{S}_T(\bj(i,k,m))) \right) - \log (K^2(T-1))  \\
  & \ge & \eta \max_{(i,k,m)}
          \wh{S}_T(\bj(i,k,m)) - \log (K^2(T-1))~.
\end{eqnarray*}
On the other hand, $\log(W_T/W_0)= \sum_{t=1}^T\log(W_t/W_{t-1})$ and
for each $t\in [T]$, we have
\begin{eqnarray*}
\lefteqn{  
  \log \frac{W_t}{W_{t-1}}   }\\
  & = &
             \log \left(\sum_{(i,k,m)}
                          p_{t,(i,k,m)} \exp(\eta \wh{Y}_t(i,k,m)) \right)  \\
& = & 2\eta +             \log \left(\sum_{(i,k,m)}
                          p_{t,(i,k,m)} \exp(- \eta  \wh\ell _{t, (i,k,m)} ) \right)  \\
& \le & 2\eta +             \log \left(\sum_{(i,k,m)}
                          p_{t,(i,k,m)} \left(1 - \eta \wh\ell
        _{t,(i,k,m)} + \frac{\eta^2}{2}  \wh\ell _{t, (i,k,m)}^2
        \right) \right)  \\
& & \qquad \text{(using $e^{-x} \le 1-x +x^2/2$ for $x\ge 0$)}   \\
& \le & 2\eta 
                          - \eta \sum_{(i,k,m)}
        p_{t,(i,k,m)}\wh\ell _{t, (i,k,m)}  +
        \frac{2\eta^2K}{\epsilon} \sum_{(i,k,m)}  p_{t,(i,k,m)}  \wh\ell _{t, (i,k,m)} 
\\
& & \qquad \text{(using $\log(1+x) \le x$ for all $x\ge -1$ and
    $\wh\ell _{t, (i,k,m)} \le 2K/\epsilon$)}   \\
& \le &                            \eta \sum_{(i,k,m)}
        p_{t,(i,k,m)}\wh{Y}_{t, (i,k,m)} +
        \frac{2\eta^2K}{\epsilon} \sum_{(i,k,m)}  p_{t,(i,k,m)}  \wh\ell _{t, (i,k,m)} \\
& & \qquad \text{(by definition of $ \wh\ell _{s, (i,k,m)} =2-\wh{Y}_{s, (i,k,m)}$)}     
\end{eqnarray*}
Comparing the upper and lower bounds obtained for $\log(W_T/W_0)$, we
have that for all $(i',k',m')\in [K]\times [K]\times [T-1]$, 
\begin{eqnarray*}
   \sum_{t=1}^T\sum_{(i,k,m)}
        p_{t,(i,k,m)}\wh{Y}_{t, (i,k,m)} & \ge &
                                                 \wh{S}_T(\bj(i',k',m'))    \\
& &        - \frac{\log (K^2(T-1))}{\eta} - \frac{2\eta K}{\epsilon}\sum_{t=1}^T\sum_{(i,k,m)}  p_{t,(i,k,m)}  \wh\ell _{t, (i,k,m)} ~.
\end{eqnarray*}
Taking expected values on both sides and noting that       
\[
    \EXP   \sum_{t=1}^T \sum_{(i,k,m)} p_{t,(i,k,m)} Y_{t, (i,k,m)} = \EXP
       \ol{S}_T~,
     \]
we get 
\[
\EXP \ol{S}_T \ge \max_{(i,k,m)} S_T(\bj(i,k,m)) - \frac{\log
  (K^2(T-1))}{\eta} - \frac{4T\eta K}{\epsilon}~,
\]
and therefore
\[
  \EXP \Reg_T \le   \frac{\log
  (K^2(T-1))}{\eta} + \frac{4T\eta K}{\epsilon} +   2T\epsilon~.
\]
Optimizing the upper bound suggests the choices
\[
  \epsilon=K\sqrt{\eta}  \quad \text{and}
  \quad \eta= \left(\frac{\log(K^2(T-1))}{2TK}\right)^{2/3}~,
\]
yielding the announced upper bound.
\end{proof}

\section{Subscription model: coupon rewards}
We finally consider the subscription model with so-called ``coupons'' fidelity rewards. Here the player earns an additional reward $r_j \in [0,1]$ after each consecutive $\rho_j \in \mathbb{N}$ plays of an arm $j$. As we will see, this setting is similar to the loyalty points model with coupon rewards which was studied in Section~\ref{sec:loycoup}. However, a key difference is that here, in order to get any additional fidelity reward, an arm needs to be played $\rho_j$ times \emph{consecutively}.

\subsection{Minimal sufficient sets}
We begin by considering the class of minimal sufficient sets for the subscription model with coupon rewards. For this, we note that due to the need to be playing an arm continuously in order to receive any fidelity reward, any optimal sequence $\bj \in [K]^T$ will mainly consist of consecutive instances of any arm, where the length of these segments is determined by $\rho_j$, the period of arm $j$. Moreover, it can easily be seen that if $\rho_j=\rho$ for all arms $j$ and $T$ is divisible by $\rho$, then the optimal strategy is to play the arm maximizing $\mu_j + r_j/\rho$ for all rounds $t=1,\dots, T$, and thus the minimal sufficient sets are the single arm strategies $\J_0$. In the more general case, the minimal sufficient sets may not simply be single arm strategies. However, by an argument similar to the one in the loyalty points setting, we can show that the total reward accumulated by the best single arm strategy is not too far from that of the best minimal sufficient set. In particular,
\begin{lemma} \label{lem:rhobound}
Let $\ol{\rho}$ denote the least common multiple of the period lengths $\rho_1,\ldots,\rho_K$. Then,
\[ \max_{\bj \in \cup_{\J \in \cC_f} \J} S_T(\bj) - \max_{\bj' \in \J_0} S_t(\bj') \leq 2\ol{\rho}~. \]
\end{lemma}

Therefore, as in the coupons loyalty points model, we compare our algorithms to single arm strategies knowing that the maximal reward of such strategies does not differ too much from the true maximal reward.

\subsection{Stochastic rewards}
\label{sec:stoccoupsub}
From the above discussion, it is clear that to gain the additional fidelity reward, each time an arm $j$ is played, it should be played for $\rho_j$ rounds consecutively. Since the fidelity functions are known, so are the  $\rho_j$.
Therefore, we consider algorithms that run in batches, that is, they select an arm $j$ using all the available data up to that point, then play that arm $\rho_j$ times before updating the estimates and selecting another arm. While in principle, any standard algorithm for the multi-armed bandit problem could be used to select the arms at the beginning of each batch, for simplicity, here we consider a batch version of the UCB algorithm \citep{auer2002finite}. We aim to accumulate rewards comparable to $T \max_{j \in [K]} (\mu_j + r_j/\rho_j)$ and so we use confidence bounds of the form.

\begin{align}
	\UCB_t(j) =  \ol X_{t,j} + \frac{r_j}{\rho_j} + \sqrt{\frac{2\log(KT)}{N_{t,j}}}, \label{eqn:cbsubcoup}
\end{align}
where $\ol X_{t,j}$ is the average base reward from $N_{t,j}$ plays of arm $j$ up to time $t$.
Note that each arm $j$ is played $\rho_j$ times when it is selected.
This leads to the following regret bound, the complete proof of which is given in Appendix~\ref{app:regBUCBcoupsub}.
\begin{theorem} \label{thm:stoccoupsub}
The expected regret of the Batch-UCB algorithm up to horizon $T$ in the stochastic coupons subscription model can be bounded by
 \[ \E[\Reg_T] \leq  \sum_{j \neq j^*} \frac{16\log(TK)}{\wt \Delta_j} + \sum_{j \neq j^*} \rho_j\wt \Delta_j + (1+\frac{2}{K})\sum_{j \neq j^*} \wt \Delta_j + 2 \ol{\rho} \]
 where $\wt \Delta_j = \max_{i \in [K]}(\mu_i + r_i/\rho_i)  - (\mu_j + r_j/\rho_j)$. 
 So the worst case regret bound is $ \E[ \Reg_T] = O (\sqrt{KT\log(T)} + \sum_{j =1}^K \rho_j  + \ol{\rho})$.
\end{theorem}

\subsection{Adversarial rewards}
\label{sec:advcoupsub}
Since we know that the best single arm strategy is almost optimal in the subscription coupons model, we aim to develop an algorithm that is competitive with that. Unlike in the loyalty points model with coupon rewards where we could simply use a standard adversarial bandits algorithm with augmented rewards, the subscription structure of the fidelity rewards here means that we need to be a bit more careful to guarantee that the augmented rewards are actually similar to the accumulated rewards. Specifically, this means that we need to play in batches and play each arm $j$ some multiple of $\rho_j$ times whenever it is selected.
For example, we may select arms using the  EXP3 \citep{auer2002nonstochastic} algorithm and then play them $\ol{\rho}$ times, where $\ol{\rho}$ is again the least common multiple of the period lengths $\rho_1,\dots, \rho_K$. This gives the following regret bound,

\begin{theorem} \label{thm:advcoupsub}
Selecting arms according to the EXP3 algorithm with parameter $\lambda = \min \{1,$ $\sqrt{\frac{\ol{ \rho} K \log(K)}{(e-1)T}}\} $ and playing each arm $j$ $\ol\rho$ times when selected in the adversarial coupons subscription fidelity bandits model leads to regret of
\[ \E[\Reg_T^\sharp] =\wt O( \sqrt{\ol{\rho} KT \log(K)})~,
\]
where $\ol{\rho}$ is again the least common multiple of the period lengths $\rho_1,\dots, \rho_K$.
\end{theorem}
\proof
The proof follows by considering an adversarial bandits problem over $S = \lfloor \frac{T}{\ol{\rho}} \rfloor$ rounds. In each round $s=1, \dots, S$, the algorithm selects an arm $J_s$ using the EXP3 algorithm and receives reward $\frac{\sum_{t \in T_s} X_{t,j} + r_j}{ 2\ol{\rho} } \in [0,1]$ where $T_s$ is such that $|T_s|= \ol{\rho}$ and represents the set of time points $t=\ol{\rho}s +1, \dots, \ol{\rho}(s+1)$ in the fidelity bandits problem where we are playing arm $j$ $\ol{\rho}$ times consecutively. Let $\ol{\Reg}_S$ denote the regret in this modified problem compared to playing arm $j^* = \argmax_{1\leq j \leq K} \frac{\sum_{t=1}^T X_{t,j} + S r_j}{\ol{\rho}}$, which we note is the same arm as the best single arm strategy in the original problem. Moreover, by Lemma~\ref{lem:rhobound}, $\Reg_T \leq 2\ol{\rho} \ol{\Reg}_S + \ol{\rho}$, so it suffices to provide a bound on $\ol{\Reg}_S$. This can be done by applying the analysis of EXP3 in Corollary~3.2 of \citep{auer2002nonstochastic}. In particular, for parameter $\lambda = \min \{1,  \sqrt{\frac{\ol{\rho} K \log(K)}{(e-1)T}}\} $, $\E[\ol{\Reg}_S] = O(\sqrt{\frac{KT\log(K)}{\ol{\rho}}})$, thus giving the result.
\endproof

\begin{remark}
Note that when $\rho_1=\cdots =\rho_K=\rho$, then the regret is simply $\wt O(\sqrt{\rho KT})$. However, in other cases, depending on the periodicities, $\ol{\rho}$ can be significantly larger than any single $\rho_j$. By a simple modification of the argument, the regret bound can be improved to $\wt O( \frac{\max_j \rho_j}{\sqrt{\min_j \rho_j}} \sqrt{KT})$ by considering a variant of EXP3 that plays each arm $j$ exactly $\rho_j$ times when it is selected. However, to avoid some technicalities arising from the fact that the number of rounds becomes random, we chose to present the simplest version. 
\end{remark}

\begin{remark}
  The above result assumes that $\rho_j$ is constant for all $j \in [K]$. Clearly if $\rho_j$ are such that $\ol{\rho}= \Omega(T)$, the bound in Theorem~\ref{thm:advcoupsub} becomes vacuous. In such a case,
  one may simply ignore the fidelity reward of any arm with $\rho_j = \Omega(T)$ as it is not received often enough to make a significant difference to the cumulative reward. In fact, for $\rho_j=\Omega(T^{1/4})$ it is better to ignore the fidelity rewards. We omit the straightforward details.
\end{remark}

\begin{remark}
From Theorems~\ref{thm:stoccoupsub}~and~\ref{thm:advcoupsub}, we observe that the cost of playing in batches is additive in the stochastic case, and multiplicative in the adversarial setting. This is to be expected since playing in batches can almost be viewed as a form of delayed feedback. It is known that the penalty for receiving delayed feedback  is multiplicative in the adversarial setting (see, e.g., \citet{joulani2013online}), so our results are consistent with this.%
\end{remark}

\section{Conclusion}
In this paper we have studied several instances of the fidelity bandits problem where the reward of each arm is augmented by a fidelity reward which measure how loyal the player has been to that arm in the past.
Our focus has been on the settings with adversarially generated base rewards, although we also analyze the stochastic version of the problem for completeness. 
 For this problem, the definition of the regret when the rewards are
 adversarial is nontrivial. By considering the stochastic analogue of
 the problem we suggest several natural regret definitions. In
 particular, we define the regret with respect to a class of policies
 which may be optimal for some configuration of the stochastic
 problem, a technique which we believe may be applicable in many
 settings beyond the fidelity bandits problem studied here. Our main
 interest in this paper was to determine when it is possible for these
 regrets to be sub-linear. We considered two forms of fidelity reward,
 namely the loyalty points model where the fidelity reward depends on
 the number of times of an arm was previously played, and the subscription model where the fidelity reward is a function of the number of consecutive plays of an arm up to that point. 

 Our findings show that learning with adversarial rewards and
 increasing fidelity functions is hard, with $\Omega(T)$ bounds
 presented for even the weakest regret definition in both the loyalty
 points and subscription model.  However, these results are worst
 case, so it remains a possibility that for specific fidelity
 functions sublinear regret could be achieved (see, e.g., the results
 for $m$-step functions in Section~\ref{sec:advincsub}).  For
 decreasing fidelity rewards, the picture is more positive. Although
 we provide $\Omega(T)$ lower bounds for the strongest notions of
 regret in the loyalty points model, we show that for weaker notions
 of regret, it is possible to achieve sub-linear regret in this
 case. This means that we are able to perform comparably to some
 near-optimal sequence of actions. For the subscription model, we show
 that sublinear regret is possible for even the strong regret when the
 fidelity rewards are decreasing. The reason for this distinction is
 that in the subscription model, due to the fact that the fidelity
 reward depends on the number of cumulative plays, there is only one
 minimal sufficient set, simplifying the class of comparator policies
 for the strong regret. Lastly, we note that in the coupons rewards
 setting, where a bonus reward is obtained every $\rho_j$
 (consecutive) plays of arm $j$, it is possible to get sub-linear
 strong regret in both the loyalty points and subscription models with
 adversarial base rewards. Interestingly, in this case, the effects of
 the fidelity rewards are different in the subscription and loyalty
 points models, with the regret scaling mutliplicatively with the
 periodicity of the coupon function in the subscription case, while it
 only increases additively in the loyalty points model.

 Although we have considered several forms of fidelity reward in this
 paper, there are still other realistic assumptions one could make on
 the fidelity reward. For example, it may be interesting to consider
 more general periodic fidelity functions, or other measures of
 fidelity such as the number of times the arm has been played in the
 last $m$ plays. This is left for future work.

 We also point out that throughout the paper we have assumed that the
 entire fidelity functions are known to the forecaster, although
 typically the algorithms only require knowledge of the cumulative
 fidelity reward $F_j(T)$. It remains to be seen if this can be
 reduced to requiring no knowledge of the fidelity reward. A related
 problem is whether one can remove the need to know the horizon $T$ and
 develop anytime algorithms for fidelity bandits problems. A key
 challenge here is that the horizon is typically also used to define
 $F_j(T)$. By assuming that the average fidelity reward is stationary
 over time, one may be able to adapt our results to the anytime
 setting, but we leave this to future work.

 Lastly, we note that our aim in this paper was to understand when it
 is possible to achieve sub-linear regret in the fidelity bandits
 problem. As such, some constants or rates (particularly in the
 stochastic reward case) may not be tight. Therefore, tightening them
 and/or providing matching lower bounds remains an interesting open
 problem.

\section*{Acknowledgments}
{G. Lugosi was supported by
  by the Spanish Ministry of Economy and
  Competitiveness, Grant PGC2018-101643-B-I00 and FEDER, EU
  and by ``Google Focused Award Algorithms and Learning for AI''.
  This work was started while C. Pike-Burke was at Universitat Pompeu Fabra.
  }

\bibliographystyle{apalike}
\bibliography{../fidelity_refs,../biblio-fb}

 \appendix

\section{Proofs for the loyalty points model with increasing rewards} \label{app:inc}
\subsection{Proof of Lemma~\ref{lem:loyinc} (Optimality of single arm strategies for increasing loyalty points bandits)} \label{app:loyincopt}
\proof
Suppose that a sequence $\bj$ plays arm $j$  
$N_j$ times up to time $T$ (and therefore $\sum_{j=1}^K N_j=T$).
Then the total (pseudo) reward of such a sequence is 
\begin{eqnarray*}
\wt S_T(\bj) = \sum_{j=1}^K \left(N_j \mu_j + \sum_{n=1}^{N_j} f_j(n) \right) 
& = &
\sum_{j=1}^K N_j \left(\mu_j + \frac{1}{N_j} F_j(N_j) \right) \\
& \le & 
\sum_{j=1}^K N_j \left(\mu_j + \frac{1}{T} F_j(T) \right) \\
& & \text{(since the $f_j$ are increasing functions)} \\
& \le & T \max_{1 \leq j \leq K} \left\{\mu_j  + \frac{1}{T} F_j(T)\right\} = \wt S_T(j^*,j^*,\ldots,j^*)~. 
\end{eqnarray*}
Hence there is a  single-arm strategy whose total reward is at least as large as that of $\bj$.

\endproof

\subsection{Proof of Theorem~\ref{thm:regstocsubinc} (Regret of UCB in stochastic loyalty points bandits with increasing fidelity rewards)} \label{app:regstocsubinc}
\proof
This proof is similar to the standard proofs of the regret bounds of
UCB (see \citet{auer2002finite,LaSz20}) 
with some adjustments to deal with the fidelity functions.

Let $B = \cap_{j=1}^K \cap_{t=1}^T \{\mu_j  + F_j(T)/T \leq \UCB_t(j)\} $ be the event that the upper confidence bounds on each arm hold at all time steps. We show that on event $B$, the number of plays of sub-optimal arms is small. First, we write
\begin{eqnarray*}
  \Reg_T 
&=& \sum_{t=1}^T (\mu_{j^*} - \mu_{J_t} + f_{j^*}(t) - f_{J_t}(N_{t-1,J_t}))) \\
& = & \sum_{j=1}^K N_{T,j} (\mu_{j^*} - \mu_j) + \sum_{t=1}^T f_{j^*}(t) - \sum_{j=1}^K \sum_{n=1}^{N_{T,j}} f_j(n)
\\ 
& = & \sum_{j=1}^K N_{T,j} (\mu_{j^*} - \mu_j) + \sum_{n=N_{T,j^*}}^T f_{j^*}(n) - \sum_{j:j\neq j^*} \sum_{n=1}^{N_{T,j}} f_j(n)
\\ 
& \leq & \sum_{j=1}^K N_{T,j} (\mu_{j^*} - \mu_j)  + (T-(T-\sum_{j:j\neq j^*} N_{T,j})) f_{j^*}(T) - \sum_{j: j \neq j^*} N_{T,j} f_j(0)
\\ & & \text{(since the fidelity functions are increasing)} 
\\ 
& \leq & \sum_{j: j \neq j^*} N_{T,j} (\mu_{j^*} - \mu_j + f_{j^*}(T) - f_j(0))~,
\end{eqnarray*}
since $N_{T,j^*} = T-\sum_{j \neq j^*} N_{T,j}$.

Hence, it suffices to bound the number of plays of each sub-optimal arm. Let $m_j = \frac{8\log(KT)}{\widetilde \Delta_j^2}$ and observe that on the event $B$, if $N_{t,j}>m_j$ arm $j$ will not be played again. Indeed, in this case, for any number of plays $m^*$ of the optimal arm,
\begin{align*}
\UCB_t(j) &= \ol{X}_{t,j} + \frac{1}{T} F_j(T) + \sqrt{ \frac{2
            \log(KT)}{m_j}} \leq \mu_j + \frac{1}{T} F_j(T) + 2\sqrt{ \frac{2 \log(KT)}{m_j}} 
\\ &< \mu_j + \frac{1}{T} F_j(T) +\widetilde  \Delta_j  = \mu_{j^*} +
     \frac{1}{T} F_{j^*}(T) \leq \UCB_t(j^*)
\end{align*}
and therefore arm $j$ is not played again.

Then, using the fact that the confidence bounds hold with probability
$1/(KT)^2$ (by Hoeffding's inequality), 
the expected regret may be bounded by
\begin{eqnarray*}
\E \Reg_T & = & \E[\Reg_T \one_{ B}] + \E[\Reg_T \one_{B^C}]
\\ 
& \leq & \sum_{j: j \neq j^*} \E[ N_{T,j} \one_{B }] (\mu_{j^*} - \mu_j +f_{j^*}(T) - f_j(0)) + T \PROB(B^C)
\\ 
& \leq & \sum_{j: j \neq j^*} \frac{8\log(KT)  (\mu_{j^*} - \mu_j +f_{j^*}(T) - f_j(0))}{\widetilde \Delta_j^2} + T \sum_{t=1}^T \sum_{j=1}^K \PROB(\mu_j + F_j(T)/T >\UCB_t(j))
\\  
& \leq  & \sum_{j: j \neq j^*} \frac{8\log(KT)  (\mu_{j^*} - \mu_j +f_{j^*}(T) - f_j(0))}{\widetilde \Delta_j^2} + \frac{1}{K}~,
\end{eqnarray*}
proving the result. To obtain the worst case regret bound, we use standard techniques (see, e.g. \cite{LaSz20}) and set $\wt \Delta_j \approx \sqrt{\frac{K \log(T)}{T}}$.
\endproof

\subsection{Proof of Theorem~\ref{thm:TregLoy} (Lower bound on the regret for the adversarial loyalty points bandit problem with increasing fidelity reward)} \label{app:loyinclb}
\proof
It suffices to consider the two-armed setting (i.e., $K=2$). The argument  may easily be extended to multiple arms, for example,
by replicating the rewards of the two arms. We may also assume that $T$ is even and the fidelity functions of each arm are the same.
We write $F(t)=\sum_{s=1}^t f(s)$ for the cumulative fidelity reward.

In order to prove the theorem, we consider just two possible sequences of rewards. 
In both cases arm $1$ has reward $X_{1,t}=\delta/5$ for all $t\in [T]$.
Also, in both cases arm $2$ has reward $X_{2,t}=0$ for all $t\in [T/2]$.
The reward $X_{2,t}$ of arm $2$ for all $t>T/2$ equals $0$ in case $(I)$, while it equals $\delta$ in case $(II)$. 

We know that in this problem single arm strategies are optimal. In case $(I)$, the optimal arm is arm $1$ with total reward $T\delta/5 + F(T)$,
while in case $(II)$, the optimal arm is arm $2$ with total reward $T\delta/2 + F(T)$.

Consider any policy and denote by $\tau_1$ the number of times the policy chooses arm $1$ up to time $T/2$.
Similarly, $\tau_2$ denotes the number of times the policy chooses arm $1$ between times $T/2+1$ and $T$.
Note that $\tau_1$ is the same in cases $(I)$ and $(II)$ as up to time $T/2$ the two cases are identical.

If $\tau_1 \le T/4$, then in case $(I)$ the policy suffers regret at least $T\delta/20$, so in the rest of 
the proof we may assume that $\tau_1 > T/4$ and we consider case $(II)$.

In that case the total reward of the policy equals
\[
   \frac{(\tau_1+\tau_2)\delta}{5}+ \frac{T\delta}{2} - \tau_2\delta + F(\tau_1+\tau_2) + F(T-\tau_1-\tau_2)~.
\]
Thus, since arm $2$ is optimal, the regret becomes
\[
 \Reg_T = \frac{- \tau_1\delta}{5} +   \frac{4\tau_2\delta}{5} + F(T) - \left(F(\tau_1+\tau_2) + F(T-\tau_1-\tau_2)\right)~.
\]
If $\tau_2>\tau_1-T/8$, then the regret is at least
\begin{eqnarray*}
   \Reg_T & \ge &
\frac{- \tau_1\delta}{5} +   \frac{4\tau_2\delta}{5}   
\quad \text{(since the fidelity reward is nondecreasing)}  \\
& \ge & \frac{3\tau_1\delta}{5} - \frac{T\delta}{10}    \\
& \ge & \frac{T\delta}{20} \quad \text{(since $\tau_1> T/4$).}  \\
\end{eqnarray*}
Finally, if $\tau_2 \le \tau_1-T/8$, then
\[
    \tau_1+\tau_2 \in \left[\frac{T}{4}, \frac{7T}{8} \right]~,
\]
and therefore, by the nondecreasing property of the fidelity rewards, for any $t \in [T]$, $F(t)+ F(T-t)$ is nonincreasing in $t \leq T/2$ and nondecreasing in $t \geq T/2$. Hence,
\begin{eqnarray*}
    F(T) - \left(F(\tau_1+\tau_2) + F(T-\tau_1-\tau_2)\right)  
& \ge & F(T) - F(7T/8) - F(T/8)    \\
    &   =  & \sum_{t=1}^{T/8}\left(f(T-t)-f(t)\right) \\
& \ge & \frac{T}{8} (f(7T/8)-f(T/8)) = \frac{T\delta}{8}~.
\end{eqnarray*}
Hence, the regret satisfies
\[
     \Reg_T \ge \frac{- \tau_1\delta}{5} +  F(T) - \left(F(\tau_1+\tau_2) + F(T-\tau_1-\tau_2)\right)
      \ge \frac{-T\delta}{10}  + \frac{T\delta}{8} = \frac{T\delta}{40}~. 
\]
\endproof

\section{Results for Stochastic Coupons Subscription Model} \label{app:coupsub}
\subsection{Proof of Theorem~\ref{thm:stoccoupsub} (Regret bound for the Batch-UCB algorithm in the stochastic coupons subscription model)} \label{app:regBUCBcoupsub}
We first bound the number of plays of any sub-optimal arm.
\begin{lemma}\label{lem:nbound}
For any sub-optimal arm $j$,
\[ 0 \leq \E[N_j(T)] \leq \frac{16\log(TK)}{\wt \Delta_j^2} + \rho_j +1 + \frac{2}{K} \]
\end{lemma}
\proof
Let $B$ be the event that the confidence bounds hold for all arms at all time steps, that is, $B= \cap_{j=1}^K \cap_{t=1}^T \{\mu_j \leq \UCB_t(j)\} $. Then define $n^*_j = \lceil \frac{16\log(TK)}{\wt \Delta_j^2} \rceil = m^*_j \rho_j + l_j$ for some integer $m^*_j \in \N$ and remainder $0 \leq l_j\leq \rho_j$. We now show that on event $B$, if we have played arm $j$ $\wt n^*_j = (m^*_j+1)\rho_j \geq n^*_j$ times, then it is not played again. Indeed, if $t$ is such that $N_{j,t} \geq \wt n_j^* \geq n^*_j$, then,
\begin{align*}
\UCB_t(j) = \ol Y_{j,t} +2 \sqrt{\frac{\log(TK)}{N_{T,j}}} \leq \wt \mu_j +  4\sqrt{\frac{\log(TK)}{n^*_j}} <\wt \mu_j + \wt \Delta_j = \wt \mu_{j^*} \leq \UCB_t(j^*).
\end{align*}
Here we have used the definition of the confidence bounds and the notation $\wt \mu_j = \mu_j + \frac{r_j}{\rho_j}$ and $j^* = \argmax_{1 \leq j \leq K}\mu_j + \frac{r_j}{\rho_j}$ .
Hence, $\UCB_t(j) < \UCB_t(j^*)$ and so we would play arm $j^*$ and not play arm $j$ again. Consequently arm $j$ is not played more than $n^*_j$ times.

Using this, we bound the expected number of times we play arm $j$ as,
\begin{align*}
\E[N_{T,j}] &= \E[N_{T,j} \one_{\{B\}}] + \E[N_{T,j} \one_{\{B^C\}}] \leq n_j^*+ T \PROB(B^C)
\\ & \leq (m^*_j+1)\rho_j + T \sum_{j=1}^K \sum_{t=1}^T \PROB(\mu_j> \UCB_t(j)) \leq \bigg \lceil \frac{16\log(TK)}{\wt \Delta_j^2} \bigg \rceil  + \rho_j + \frac{2}{K}
\end{align*}
where we have used the fact that the confidence intervals hold with probability at least $1-\frac{2}{T^2K^2}$. Using the trivial bound $\lceil x \rceil \leq x +1$ and noting that $N_j(T) \geq 0$ gives the result of the lemma.
\endproof

Using this, we now prove the regret bound.
\proof
We play each arm $j$ in batches of $\rho_j$ plays. Denote the number of such batches of arm $j$ as $B_{T,j}$ and note that $N_{T,j} = \rho_j B_{T,j}$. Using Lemma~\ref{lem:rhobound}, we write the regret as,
\begin{align*}
\E[\Reg_T] &= T(\mu_{j^*} + \frac{r_{j^*}}{\rho_{j^*}}) + 2 \ol{\rho} - \sum_{j=1}^K\E[B_{T,j}](\rho_j \mu_j + r_j) 
\\ &= T(\mu_{j^*} + \frac{r_{j^*}}{\rho_{j^*}}) - \sum_{j=1}^K \E[N_{T,j}] (\mu_j + \frac{r_j}{\rho_j}) + 2 \ol{\rho}
=  \sum_{j \neq j^*} \E[N_{T,j}] \wt \Delta_j + 2 \ol{\rho}
\end{align*}
where $\wt \Delta_j = \max_{1 \leq j \leq K} \{ \mu_j + \frac{r_j}{\rho_j}\} - ( \mu_j + \frac{r_j}{\rho_j}) $. %
Substituting the result from Lemma~\ref{lem:nbound} into the above result gives,
\begin{align*}
\E[\Reg_T] %
\leq \sum_{j \neq j^*} \frac{16\log(TK)}{\wt \Delta_j} + \sum_{j \neq j^*} \rho_j\wt \Delta_j + (1+\frac{2}{K})\sum_{j \neq j^*} \wt \Delta_j + 2 \ol{\rho}.
\end{align*}
\endproof

\end{document}